\documentclass[a4paper,fleqn]{cas-dc}

\usepackage[numbers]{natbib}

\def\tsc#1{\csdef{#1}{\textsc{\lowercase{#1}}\xspace}}
\tsc{WGM}
\tsc{QE}
\tsc{EP}
\tsc{PMS}
\tsc{BEC}
\tsc{DE}
\usepackage{amsthm}

\newtheorem{theorem}{Theorem}

\newtheorem*{theorem*}{Theorem}
\newtheorem{remark}{Remark}%
\newtheorem*{corollary*}{Corollary}
\newtheorem*{assumption*}{Assumption}
\newtheorem{definition}{Definition}%
\newtheorem{assumption}{Assumption}
\newtheorem{corollary}{Corollary}

\usepackage{algorithm}
\usepackage{algorithmicx}%
\usepackage{algpseudocode}%

\usepackage{subfigure}
\usepackage{multicol}
\usepackage{booktabs}
\usepackage{hyperref}

\usepackage{graphicx}

\begin{document}
\let\WriteBookmarks\relax
\def\floatpagepagefraction{1}
\def\textpagefraction{.001}
\shorttitle{PACER: A Fully Push-forward-based Distributional RL Algorithm}
\shortauthors{Wensong Bai et~al.}

\title [mode = title]{PACER: A Fully Push-forward-based Distributional Reinforcement Learning Algorithm}                      

\author[1]{Wensong Bai}
\ead{wensongb@zju.edu.cn}

\credit{Conceptualization of this study, Methodology, Software} 

\affiliation[1]{organization={College of Computer Science and Technology},
                addressline={Zhejiang University}, 
                city={Hangzhou},
                postcode={310063}, 
                state={Zhejiang},
                country={China}}

\author[1,2]{Chao Zhang}
\cormark[1]
\ead{zczju@zju.edu.cn}

\author[1]{Yichao Fu}
\ead{fuyichao@zju.edu.cn}

\author[3]{Peilin Zhao}
\ead{masonzhao@tencent.com}

\author[1]{Hui Qian}
\ead{qianhui@zju.edu.cn}

\author[1]{Bin Dai}
\ead{bindai@zju.edu.cn}

\affiliation[2]{organization={Advanced Technology Institute},
                addressline={Zhejiang University}, 
                city={Hangzhou},
                postcode={310063}, 
                state={Zhejiang},
                country={China}}

\affiliation[3]{organization={Tencent AI Lab},
                city={Shenzhen},
                postcode={518000}, 
                state={Guangdong},
                country={China}}

\nonumnote{This work was supported in part by the National Natural Science Foundation of China: 62206248.}
\cortext[cor1]{Corresponding author}


\begin{abstract}
    In this paper, we propose the first fully push-forward-based distributional reinforcement learning algorithm, named PACER, which consists of a distributional critic, a stochastic actor and a sample-based encourager.
    Specifically, the push-forward operator is leveraged in both the critic and actor to model the return distributions and stochastic policies respectively, enabling them with equal modeling capability and thus enhancing the synergetic performance. Since it is infeasible to obtain the density function of the push-forward policies, novel sample-based regularizers are integrated in the encourager to incentivize efficient exploration and alleviate the risk of trapping into local optima. Moreover, a sample-based stochastic utility value policy gradient is established for the push-forward policy update, which circumvents the explicit demand of the policy density function in existing REINFORCE-based stochastic policy gradient. As a result, PACER fully utilizes the modeling capability of the push-forward operator and is able to explore a broader class of the policy space, compared with limited policy classes used in existing distributional actor critic algorithms (i.e. Gaussians). We validate the critical role of each component in our algorithm with extensive empirical studies. Experimental results demonstrate the superiority of our algorithm over the state-of-the-art.
\end{abstract}




\maketitle

\section{Introduction}
Distributional reinforcement learning (DRL) considers the intrinsic randomness of returns by modeling the full distribution of discounted cumulative rewards \citep{bellemare2017distributional}.  In contrast to their counterparts that solely model the expected returns, the skewness, kurtosis, and multimodality of return distributions can be carefully captured by DRL algorithms, usually resulting in more stable learning process and better performance \citep{lyle2019comparative}. The SOTA has been achieved by DRL algorithms in various sequential decision-making and continuous control tasks \citep{yue2020implicit}. 

Recently, the thriving of DRL has catalyzed a large body of algorithmic studies under the \emph{actor-critic} framework which leverage \emph{push-forward} operator to parameterize the return distribution in the critic step \citep{ma2020dsac,dabney2018distributional,choi2021risk,duan2021distributional}.
Actually, the push-forward idea, which has played an important role in optimal transport theory \citep{villani2009optimal} and recent Monte Carlo (MC) simulations \citep{marzouk2016sampling,parno2018transport,peherstorfer2019transport}, incarnates an efficacious approach for modeling complicated distributions through sampling. This concept plays a vital role in the distributional temporal-difference learning procedure of the critic \citep{bdr2022}. 

Nevertheless, adopting push-forward operator solely to the critic, as in existing distributional actor-critic (DAC)  algorithms, is far from sufficient to achieve optimal efficacy, due to that the critic and the actor are highly interlaced into each other. Concretely, DAC algorithms are \emph{two-time-scale} procedures in which the critic performs temporal-difference (TD) learning with an approximation architecture, and the other way around, the actor is updated in an approximate gradient direction based on information provided by the critic \citep{konda1999actor}; Thus, it is reasonable to conjecture that only by adopting highly expressive push-forward operators in both parts can the procedure ignite an enhanced performance.\footnote{Indeed, there are some alternative ways to enhance the expressiveness of policies in the literature, for example, \citep{yue2020implicit} propose to use semi-implicit mixture of Gaussians to model the policy. However, the diagonal variance simplification still hampers its modeling capability.} The observed improvement in our experiments also confirms this viewpoint, as brought about by the fully push-forward approach.


However, directly incorporating the push-forward operator to construct an actor is virtually infeasible in current DAC framework, mainly due to the following two challenges.
\begin{enumerate}
	\item \textbf{Gradient calculation.} Generally, policies equipped with push-forward operator can only generate decision samples, and it is impossible to explicitly acquire the policies' density functions. As a result, this limitation would fail the policy update procedure in existing DAC framework, since it requires the logarithmic density to construct the REINFORCE stochastic policy gradient $\mathbb{E}Q^{\pi_\theta}(s,a)\nabla \log\pi_\theta(a|s)$ \citep{heess2015learning}.
	\item \textbf{Exploration controlling.} Based on the maximum entropy principle \citep{ziebart2010modeling}, existing DAC algorithms highly rely on the entropy regularizer to encourage sufficient exploration during learning, which again demands analytic density functions \citep{ma2020dsac,yue2020implicit,choi2021risk}. Nevertheless, as push-forward policies do not have explicit density, it is not feasible to calculate their entropy $\mathcal{H}\big(\pi_\theta(\cdot|s) \big)$.
\end{enumerate} 

To bridge this gap, we propose a practical fully push-forward-based DRL algorithm, named Push-forward-based Actor-Critic-EncourageR (PACER), which applies reparameterization technique in policy gradient calculation and utilizes sample-based regularizer to encourage exploration. 
PACER incorporates three key ingredients: 
(\romannumeral1) An \emph{actor} makes decisions according to a \emph{push-forward} policy transformed from a basis distribution by deep neural networks (DNNs); (\romannumeral2) A \emph{critic} models return distributions using \emph{push-forward} operator and evaluates the policy via employing utility functions on rewards; (\romannumeral3) An \emph{encourager} incentivizes exploration by guiding the actor to reduce a \emph{sample-based} metric, for example the maximum mean discrepancy or $p$-Wasserstein distance, between the push-forward policy and a reference policy.
We summarize our main contributions as follows.

\begin{enumerate}
	\item PACER is the first DAC algorithm that simultaneously leverages the push-forward operator in both actor and critic networks. PACER fully utilizes the modeling capability of the push-forward operator, resulting in significant performance boost.
	\item A stochastic utility value policy gradient (SUVPG) theorem is established to construct a reparameterized push-forward policy gradient. According to this theorem, the stochastic policy gradient for PACER can be readily calculated solely with decision samples. 
	\item A set of sample-based regularizers are designed to facilitate the training of push-forward policies, enabling efficacious exploration of intricate environments with stochastic returns, alleviating the occurrence of trapping into poor local optima.
\end{enumerate}

Empirical studies are conducted on several complex sequential decision-making and continuous control tasks. Experimental results demonstrate that: 
(\romannumeral 1) Push-forward policies show sufficient exploration ability and would not degenerate into deterministic policies expeditiously;
(\romannumeral 2) Push-forward policies combined with sample-based regularizers suffice to ensure superior performance;
(\romannumeral 3) PACER surpasses other baselines and achieves new SOTAs on all testing tasks.

The rest of this article is organized as follows. Section \ref{sec: related work} discusses the related works. Section \ref{sec: Preliminaries} introduces the preliminaries of the Markov decision process, distributional reinforcement learning, and implicit quantile network. Our PACER algorithm and main theorem are elaborated in Section \ref{sec: Methodology}. Empirical results are reported in Section \ref{sec: exp} to illustrate the efficiency of the proposed method, and conclusions are drawn in Section \ref{sec: conclusion}.

\section{Related Works} \label{sec: related work}
\subsection{Distributional reinforcement learning}
In the early stage of DRL, the return distribution is usually restricted to certain distribution class, such as Gaussian class or the Laplace class \cite{sato2001td,engel2005reinforcement,morimura2010parametric}. 
However, this restriction may lead to significant discrepancies between the chosen distribution class and the truth, thereby introducing substantial estimation errors during the value evaluation process \cite{duan2021distributional}.
Recently, distribution-agnostic methods are investigated in depth, trying to reduce the estimation error \cite{rowland2018analysis}. 
\citep{bellemare2017distributional} proposes a categorical representation, which utilizes the discrete distribution on a fixed support to model the random return. 
Later, quantile return representation, e.g. Quantile Regression Deep Q-Network (QRN) \cite{dabney2018distributional}, Implicit Quantile Network (IQN) \cite{dabney2018implicit}, Fully Parameterized Quantile Function (FQF) \cite{yang2019fully}, are proposed to overcome the limitation of the fixed support.
Typically, this representation leverages the \emph{push-forward} operator to dynamically adjust quantiles of the return distribution, and it reveals strong expressiveness to model any complex return distributions.
Currently, the quantile representation is the principle way to model the return distribution, which has been shown to yield low value estimation errors in various studies \cite{dabney2018distributional,dabney2018implicit,yang2019fully,yue2020implicit,ma2020dsac,choi2021risk,duan2021distributional}.

The DAC algorithms, based on a distributional version of Actor-Critic frame, have achieved the state-of-the-art performance in the DRL regime \cite{ma2020dsac,nam2021gmac,yue2020implicit,choi2021risk,duan2021distributional}.
The first DAC algorithm is the D4PG algorithm \cite{barth2018distributed}, which is a distributional version of Deep Deterministic Policy Gradient (DDPG) algorithm \cite{lillicrap2015continuous} with categorical return distribution representation. 
This method is later improved by using the quantile representation to replace the categorical representation by SDPG \cite{singh2022sample}. 
Both D4PG and SDPG manually add noise to deterministic policies to enhance the diversity of actions.
In addition to D4PG/SDPG that utilizing deterministic policies with noise, there is another category of entropy-regularized DAC algorithms known as Distributional Soft-Actor-Critic (DSAC) \cite{ma2020dsac,choi2021risk,duan2021distributional}.
DSAC algorithms leverage stochastic policies with entropy regularizers to enhance exploration \cite{ma2020dsac,kuznetsov2020controlling,choi2021risk,duan2021distributional}. 
Combined with the quantile representation \cite{dabney2018implicit}, DSAC algorithms generally outperform DAC algorithms employing deterministic policies with noise \cite{ma2020dsac,duan2021distributional}.

Despite achieving SOTA performance, existing DSAC algorithms assume Gaussian or Gaussian-like distributions for their stochastic policies, as they necessitate the computation of policy entropy and logarithmic density during training, especially in continuous action spaces. This assumption significantly constrains the range of the policy space. In contrast, PACER utilizes a sample-based regularizer to promote action diversity without imposing restrictions on the density form of the policy. Consequently, pacer can realize multimodal policy distributions during the learning procedure, bringing the benefits of more effectively utilizing return distribution information, enhancing exploration in environments, and alleviating the risk of falling into sub-optimum.

\subsection{Efficient exploration policy}
While the exploration in reinforcement learning was considered as early as 1992 \cite{thrun1992efficient}, designing effective exploration strategies remains an open challenge in the field of reinforcement learning. The most commonly employed exploration technique is random exploration, i.e., $\epsilon$-greedy \cite{sutton2018reinforcement}, which utilizes a time-decaying parameter $\epsilon$ to control the probability of taking random actions in the environment. The $\epsilon$-greedy strategy is often combined with deterministic policies, which can be categorized into value-based reinforcement learning algorithms, e.g. DQN \cite{mnih2015human}, C51 \cite{bellemare2017distributional}, QR-DQN \cite{dabney2018distributional}; and deterministic policy gradient-based algorithms, e.g. DPG \cite{silver2014deterministic}, DDPG \cite{lillicrap2015continuous}, TD3 \cite{fujimoto2018addressing}, D4PG \cite{barth2018distributed}. Although $\epsilon$-greedy methods can solve the sparse reward problem given sufficient time theoretically, it is often impractical in real-world applications due to the potentially large learning times required \cite{ladosz2022exploration}. In contrast to deterministic policy, There is rich literature aiming to obtain a high-expressive stochastic policy to encourage exploration during the training. \citep{haarnoja2017reinforcement} proposed SQL with an energy-based stochastic policy, which follows the maximum entropy principle to guarantee flexibility. In \cite{haarnoja2018soft}, SAC is proposed to mitigate the policy's expressiveness issue using reparameterization trick \cite{kingma2013auto} while retaining tractable optimization, SAC adopts a maximum entropy RL objective function to encourage exploration. 


The empirical success of the aforementioned approaches underscores the benefits of adopting well-expressive policies to effectively explore the environment. However, these methods are constrained either by the necessity of computing the logarithmic density $\nabla \log\pi_\theta(a|s)$ in SPG or by the requirement to compute the policy entropy $\mathcal{H}(\pi(\cdot|s))$ within the maximum entropy framework. These constraints typically result in the utilization of parameterized distributions, commonly Gaussian or a mixture of Gaussians, to model the policy. Employing parameterized distributions for stochastic policy significantly restricts its expressiveness, consequently impeding the agent's ability to explore the environment.
In contrast to the aforementioned methods, our proposed PACER algorithm employs a sample-based regularizer instead of entropy to incentivize exploration. Furthermore, PACER leverages a newly proposed policy gradient theorem for policy updates, eliminating the demand for policy log density by SPG. As a result, our approach substantially broadens the expressiveness of the policy, enabling more effective exploration of the environment.

\subsection{Utility functions in DRL.} 

In distributional reinforcement learning, the estimated distribution of value function models both the parametric and intrinsic uncertainties \cite{mavrin2019distributional}. \emph{Optimism in the face of uncertainty} \cite{lai1985asymptotically} is one of the fundamental exploration principles that employs parametric uncertainty to promote exploring less understood behaviors and to construct confidence set. 
Numerous studies have attempted to utilize the estimated uncertainty for exploration, such as \cite{mavrin2019distributional,ladosz2022exploration,cho2024pitfall}, and several DRL algorithms distort the learned distribution by \emph{utility functions} to create a risk-averse or risk-seeking decision-making \cite{dabney2018implicit,ma2020dsac,choi2021risk}.
Typically, there are two approaches to utilizing utility functions in DRL: (\romannumeral 1) \emph{Reward shaping} type functions, which reshape individual reward distributions to guide policy for better exploration \cite{ladosz2022exploration,lindenberg2022conjugated}; (\romannumeral 2) \emph{Risk measure} type functions, which map the whole cumulative return distribution to a real number to generate risk-sensitive policies \cite{dabney2018implicit}.
Commonly used utility functions including:
mean-variance \cite{prashanth2016variance,mavrin2019distributional,prashanth2022risk}, entropic criterions \cite{ma2020dsac,choi2021risk,duan2021distributional}, and distorted expectations \cite{balbas2009properties,chow2015risk,chow2017risk,prashanth2022risk,cho2024pitfall}. 
Albeit the selection of utility functions is highly task related, the effectiveness of leveraging utility functions in DRL algorithms has been demonstrated by various studies \cite{dabney2018implicit,mavrin2019distributional,ma2020dsac,choi2021risk,singh2020improving,cho2024pitfall}.
Among existing utility functions, the conditional value at risk (CVaR) \cite{singh2020improving} is the most widely used one, which belongs to distorted expectation family and is usually adopted to improve the robustness of DRL algorithms in the presence of risks. 

\section{Preliminaries} \label{sec: Preliminaries}
We model the agent-environment interaction by a discounted infinite-horizon Markov Decision Process (MDP) $(\mathcal{S},\mathcal{A},R,P_{r},P_{s},\mu_0,\gamma)$,
where  $\mathcal{S}$ is the state space,
$\mathcal{A}$ is the action space, and we assume they are all continuous.
$R(s,a) \sim P_{r}(\cdot|s,a)$ denotes the random reward on the state-action pair  $(s,a)$,
$P_{s}$ is the transition kernel, 
$\mu_0$ is the initial state distribution, and $\gamma \in (0,1)$ is the discounted factor.
A stationary stochastic policy $\pi(\cdot|s): \mathcal{S} \rightarrow \mathscr{P}(\mathcal{A})$ gives a probability distribution over actions based on the current state $s$. 
The state occupancy measure of $s$ w.r.t. a policy $\pi$ is defined by $d_{\mu_0}^{\pi}(s) := \sum_{t=0}^{\infty} \gamma^t P(s_t=s|\mu_0,\pi)$. The random return $Z^{\pi}(s,a) \in \mathcal{Z}$ is defined as the discounted sum of rewards $R(s_t,a_t)$ starting from $s_0 = s, a_0 = a$, i.e. $Z^{\pi}(s,a) := \sum_{t=0}^{\infty} \gamma^t R(s_t,a_t)|_{s_0 = s, a_0 = a, a_t \sim \pi}$.
Note that the classic state-action value function $Q^\pi$ is the expectation of $Z^\pi$, where the expectation takes over all sources of intrinsic randomness \cite{bdr2022}. 
While under the distributional setting, it is the random return $Z^\pi$ itself rather than its expectation that is being directly modeled. 
The cumulative distribution function (CDF) for $Z^{\pi}(s,a)$ is denoted by $F_{Z^\pi(s,a)}(z) := P(Z^{\pi}(s,a) \leq z)$, and its inverse CDF is denoted by $F_{Z^\pi(s,a)}^{-1}(\tau) := \mathrm{inf}_{z \in \mathbb{R}} \{z: F_{Z^\pi(s,a)}(z) \geqslant \tau \}$.

\subsection{Distributional Bellman equation} \label{sec: Distributional Bellman operator}
The \emph{distribution Bellman equation}  describes a recursive relation on $Z^{\pi}(s,a)$, similar as the Bellman equation on the Q function \cite{bellemare2017distributional}, 
\begin{equation}
	\label{eq: rv Bellman equation}
	Z^{\pi}(s,a) \overset{\mathcal{D}}{=} R(s,a) + \gamma Z^{\pi}(S',A'),
\end{equation}
where $\overset{\mathcal{D}}{=}$ denotes the equality in distribution. 
Based on \eqref{eq: rv Bellman equation}, a distributional Bellman operator can be constructed for the distributional TD update.
Here, we first introduce the push-forward operator, and then define the distributional Bellman operator according to it.
\begin{definition}[Push-forward operator \cite{peyre2019computational}]
	\label{push-forward-operator}
	For a continuous map $T: \mathcal{X} \rightarrow \mathcal{Y}$, we define its corresponding push-forward operator as $T_{\sharp}: \mathcal{M}({\mathcal{X}}) \rightarrow \mathcal{M}({\mathcal{Y}})$, where $\mathcal{M}(\mathcal{X})$ and $\mathcal{M}({\mathcal{Y}})$ denotes the set of probability measures on the domain $\mathcal{X}$ and ${\mathcal{Y}}$, respectively.
	Specifically, given a probability measure $\mathscr{P}_1 \in \mathcal{M}({\mathcal{X}})$, $\mathscr{P}_2 = T_{\sharp}\mathscr{P}_1$ satisfies:
	\begin{small}
		\begin{equation}
			\scalebox{0.8}{$\displaystyle\int_{\mathcal{Y}}$} h(y) \mathrm{d} \mathscr{P}_2(y) = \scalebox{0.8}{$\displaystyle\int_{\mathcal{X}}$} h(T(x)) \mathrm{d} \mathscr{P}_1(x), \forall h \in \mathcal{C}(\mathcal{Y}),
		\end{equation}
	\end{small}
	where $\mathcal{C}(\mathcal{Y})$ denotes the collection of all continuous bounded functions on $\mathcal{Y}$.
\end{definition}
Actually, the push-forward operator associated with DNNs has been widely used in the machine learning literature to approximately generate samples for complex distributions  \cite{creswell2018generative,goodfellow2020generative,ororbia2022neural}. 
Here, we use it to define the distributional Bellman operator $\mathcal{T}_d$ on a random return $Z^\pi(s,a)$.
Specifically, $\mathcal{T}_d:\mathcal{M}({\mathcal{Z}})\to\mathcal{M}({\mathcal{Z}})$  is defined as the push-forward operator associated with an affine map $f_{r,\gamma}(x) = r + \gamma x$ on $x \in \mathbb{R}$,
\begin{equation}
	\mathcal{T}_d\mathscr{P}\left({Z^\pi(s,a)}\right) = (f_{R(s,a),\gamma})_{\sharp} \mathscr{P}\left({Z^\pi(s',a')}\right), 
\end{equation}
where $s' \sim P_s(\cdot|s,a)$ and $a' \sim \pi(\cdot|s') $.
Furthermore, the contraction mapping property of the $\mathcal{T}_d$ is shown by \cite{bellemare2017distributional} when under the supreme p-Wasserstein metric $\bar{w}_p$,
\begin{equation}
\bar{w}_p\left(\mathcal{T} \mathscr{P}(Z), \mathcal{T} \mathscr{P}(Z')\right) \leq \gamma \bar{w}_p\left(\mathscr{P}(Z) , \mathscr{P}(Z')\right).
\end{equation}

\subsection{The IQN and distributional TD Learning}
Among the quantile representation of the return distributions, the implicit quantile network (IQN) \cite{dabney2018implicit} is the most widely used one in DRL algorithms.
Basically, IQN utilizes the push-forward operator to transform samples from a uniform distribution $U(0,1)$ to the corresponding quantile values that sampled from the return distributions. 
Thus, we can approximate the return distribution with an IQN-induced implicit quantile distribution, which is given as follows.

\begin{remark}[Implicit quantile distribution] 
	Given a set of sampled quantiles $\tilde{\tau} = \{\tau_1,...,\tau_{N}\} \overset{i.i.d}{\sim} U(0,1)$ that sorted by $\tau_i<\tau_{i+1}$. The implicit quantile distribution $Z_{\pi}(s,a,\tilde\tau;\theta_z)$ induced by an IQN with parameters $\theta_z$, for a random return $Z_{\pi}(s,a)$, is defined as a weighted mixture of N Diracs:
	\begin{align}\label{eqn:iqn}
		Z_{\pi}(s,a,\tilde\tau;\theta_z) := \scalebox{0.9}{$\displaystyle\sum\nolimits_{i=0}^{N-1}$}(\tau_{i+1}-\tau_i)\delta_{z(s,a,\hat{\tau}_i;\theta_z)},
	\end{align}
	where $z(s,a,\hat{\tau}_i;\theta_z) = F_{Z_{\pi}(s,a)}^{-1}(\hat{\tau}_i)$ with $\hat{\tau}_i := \frac{\tau_{i+1}+\tau_i}{2}$.
\end{remark} 
The distributional TD learning procedure can be carried out by minimizing the following Huber quantile regression loss \cite{dabney2018implicit},
\begin{small}
	\begin{equation}
		\label{eq: quantile loss}
		\rho_\tau^\kappa\left(\delta_{i j}\right) = 
		\left\{\begin{array}{ll}
			\frac{1}{2\kappa} \left|\tau-\mathbb{I}\left\{\delta_{i j}<0\right\}\right| \delta_{i j}^2, & \text { if }\left|\delta_{i j}\right| \leq \kappa , \\
			\left|\tau-\mathbb{I}\left\{\delta_{i j}<0\right\}\right| \left(\left|\delta_{i j}\right|-\frac{1}{2} \kappa\right), & \text { otherwise. }
		\end{array}\right.
	\end{equation}
\end{small}
In \eqref{eq: quantile loss}, $\kappa$ is a constant threshold, and $\delta_{i j}$ is the pairwise TD-errors between the implicit quantile approximations of two successive steps.
\begin{equation}
	\label{eq: quantile_td}
	\delta_{i j}(s,a)=r(s,a)+\gamma z (s', a',\hat{\tau}_i; \hat{\theta}_z) - z(s,a,\hat{\tau}_j;\theta_z),
\end{equation}
where  $a' \sim \pi(\cdot;s')$, $\hat{\tau}_i$ and $\hat{\tau}_j$ are calculated based on two randomly sampled quantiles $\tilde \tau'$ and $\tilde \tau$.
Note that two distinct IQNs, i.e. $\hat{\theta}_z$ and $\theta_z$, are adopted in \eqref{eq: quantile_td}, which work as the target network trick \cite{mnih2015human} that commonly used in the RL literature.

\section{Methodology} \label{sec: Methodology}


\begin{figure}[!t] 
	\centering
	\includegraphics[width=0.48\textwidth]{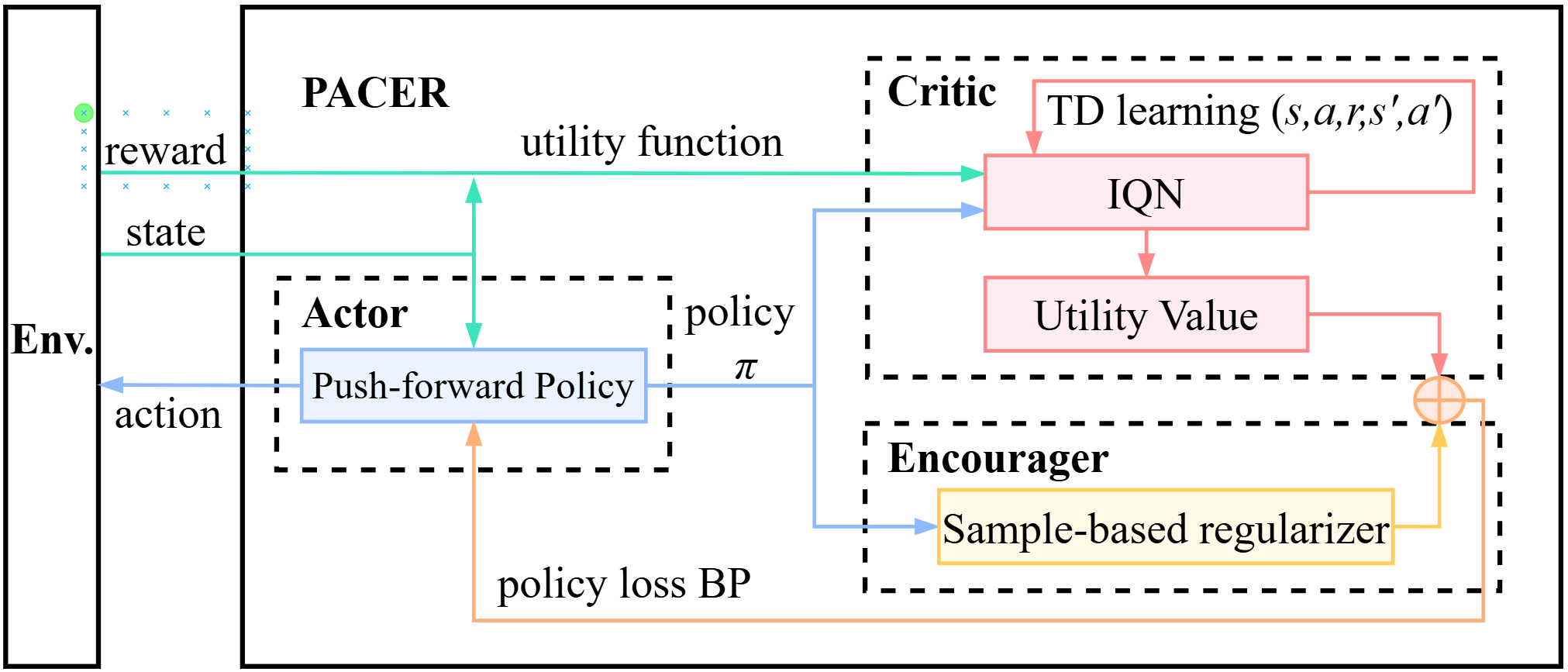} 
	\caption{The framework of PACER, where contents with identical color belong to a same module. 
		The \emph{actor} makes decisions according to a push-forward policy.
		The \emph{critic} models return distributions with an IQN and evaluates the policy via a utility function.
		The \emph{encourager} stimulates exploration by a sample-based regularizer.} \label{fig: diagram}
\end{figure}

In this section, we propose the PACER algorithm.
We first introduce the Actor-Critic-Encourager (ACE) framework of PACER.  
Then we summarize the objective function for each ingredient and establish a stochastic utility value policy gradient theorem for policy update.
The full pseudocode for PACER is given in Algorithm \ref{alg: PACER}. 

\subsection{The Actor-Critic-Encourager framework}
The framework of PACER is shown in Fig. \ref{fig: diagram}. 
It consists three main parts: an actor with push-forward policy, a critic with quantile representation, and an encourager with sample-based metric.

\subsubsection{Actor with push-forward policy}\label{sec:actor} 
The actor in PACER is a DNN which acts as a push-forward operator transforming from a base distribution $\mathscr{P}(\mathcal{X})\in\mathcal{M}(\mathbb{R}^d)$ to the action space in a \emph{sample-to-sample} manner, i.e. $a = \pi(s,\xi;\theta_\pi): \mathcal{S} \times \mathcal{X} \rightarrow \mathcal{A}$, where $\xi \sim \mathscr{P}(\mathcal{X})$ and $\theta_\pi$ are the policy network parameters. Note that we may also write $a\sim \pi(\cdot|s;\theta_\pi)$ as the push-forward policy is a stochastic policy essentially.
Practically, an action in state $s$ can be generated in a lightweight  approach by first sampling $\xi$ from $\mathscr{P}(\mathcal{X})$ and then transforming it with $\pi(s,\xi;\theta_\pi)$.
Note that this kind of push-forward distributions has been shown to have high expressiveness and modeling capability both in theory and practice \cite{baptista2023approximation,goodfellow2020generative}, and it has been widely used in machine learning literature to approximately generate samples for complex distributions \cite{kingma2014semi,creswell2018generative,goodfellow2020generative,ororbia2022neural}.  
While it is easy to obtain sample from push-forward policies,
it is generally intractable to obtain its density function explicitly.

\subsubsection{Critic with quantile return representation}
The critic utilizes an IQN to transform a sample from uniform distribution $U(0,1)$ to the corresponding quantile value sampled from the return distribution. The return distribution approximation is maintained by a weighted mixture of N Diracs. Note that there are two alternative ways, QRN \cite{dabney2018distributional} and FQF \cite{yang2019fully}, to represent quantile returns. However, QRN is designed for discrete actions, thus precluding continuous control tasks from its application.
FQF requires additional computational steps to update another network for fraction proposal. Although it can obtain benefits, the complexity it brings is not conductive to the understanding of this proposed algorithm.

We leverage the implicit quantile distribution $Z_{\pi}(s,a,\tilde\tau;\theta_z)$ defined in \eqref{eqn:iqn} to model the random return, and update it according to \eqref{eq: quantile loss}.
For better exploration or task specific risk-sensitive purposes, a utility function $\psi(\cdot)$ is adopted to reshape the policy's random reward $R(s,a)$. 
Commonly used utility functions include expectations, exponential functions, identity maps, truncated functions, and linear combinations with custom rewards that adhere to the principle of intrinsic motivation \cite{ladosz2022exploration}. By utilizing the reshaped return as the training objective, policies reflecting distinct preferences for rewards can be achieved (e.g., reward conservatism or aggressiveness). For instance, in environments with sparse rewards, the original reward signal may not provide sufficient gradients to update the policy effectively. Therefore, adopting an exponential utility function amplifies the rewards obtained through sampling, thus facilitating more targeted exploration of the environment.

Define the random return with utility as 
\begin{equation}
	Z_{\psi}^{\pi}(s,a) := \sum_{t=0}^{\infty} \gamma^t \psi(R(s_t,a_t)),
\end{equation}
where $s_0=s,a_0=a,a_{t+1} \sim \pi(\cdot|s_{t+1})$. 
The state-action utility function $Q_{\psi}^{\pi}(s,a)$ can be defined as the expectation of $Z_{\psi}^{\pi}(s,a)$:
\begin{equation} \label{eq: tate-action utility function}
	Q_{\psi}^{\pi}(s,a) := \mathbb{E}_{s_{t+1} \sim P_s, a_{t+1} \sim \pi(\cdot|s_{t+1}), R}\big[\sum_{t=0}^{\infty}\gamma^t\psi(R(s_t,a_t))\big],
\end{equation}
where $s_0=s,a_0=a$. Accordingly, the state utility function is defined by
\begin{equation}
	V_{\psi}^{\pi}(s) := \mathbb{E}_{a \sim \pi(\cdot|s)}[Q_{\psi}^{\pi}(s,a)].
\end{equation}
Correspondingly, they satisfy an iterative property analogous to the Bellman equations: 
\begin{theorem}
	\label{th: utility bellamn}
	Let $Z_{\psi}^{\pi}$ be the random return with utility of policy $\pi$, and $Q_{\psi}^{\pi}(s,a)$ be its expectation. They satisfy the following equations:
	\begin{equation}
		Z_{\psi}^{\pi} (s,a) \overset{\mathcal{D}}{=} \psi(R(s,a)) + \gamma Z_{\psi}^{\pi}(S',A'),
	\end{equation}
    \vspace{-15pt}
	\begin{equation}
		Q_{\psi}^{\pi}(s,a) = \mathbb{E}_{R}[\psi(R(s,a))] + \gamma \mathbb{E}_{s' \sim P_s} [V_{\psi}^\pi(s')].
	\end{equation}
\end{theorem}
For a given policy $\pi$, the critic can evaluate its performance with $\mathbb{E}_{s \sim \mu_0} V_{\psi}^{\pi}(s)$.


\begin{small}
	\begin{algorithm}[!t] 
		\caption{PACER: Push-forward-based Actor-Critic-EncourageR}\label{alg: PACER}
		\begin{algorithmic}[1]
			\Require environment $env$, replay buffer $\mathcal{B}$, number of quantiles K, batch size M, discounted factor $\gamma$, base distribution $\mathscr{P}(\mathcal{X})$, regularizer sample number $N_r$, utility function $\psi$, weight $\alpha$, learning rate $\beta$, etc.
			\State initialize policy networks $\theta_{\pi}$,  value networks $\theta_z$ and target networks $\hat{\theta}_z$
			\For{$n=1$ to $N$}
			\State sample a transition $\{(s_n, a_n, r_n, s'_n)\}$ from $env$
			\State $\mathcal{B} = \mathcal{B} \cup \{(s_n, a_n, \psi(r_n), s'_n)\}$
			\If{$\frac{n}{N_{update}} == 0 \  \& \  n\geq N_{explore} $}
			\State sample $\mathcal{B}_n = \{(s, a, r, s')_{m=1}^{M}\}$ from $\mathcal{B}$
			\For{each $(s, a, r, s')$ in $\mathcal{B}_n$}
			\State $\triangleright$ Distributional TD learning.
			\State generate quantiles $\{\tau_i\}_{i=1}^{K}$, $\{\tau_j\}_{j=1}^{K}$
			\State $\hat{\tau}_i = \frac{\tau_{i+1}+\tau_i}{2}$, $ \hat{\tau}_j = \frac{\tau_{j+1}+\tau_j}{2}$
			\For{$i=1$ to $K$, $j=1$ to $K$}
			\State sample $\xi \sim \mathscr{P}(\mathcal{X})$
			\State $\hat{z} = r(s,a)+\gamma z (s', \pi(s',\xi;\theta_\pi),\hat{\tau}_i; \hat{\theta}_z)$
			\State $\delta_{i j}(s,a)=\hat{z} - z(s,a,\hat{\tau}_j;\theta_z)$ 
			\EndFor
			\State $\mathcal{L}^m(\theta_z) = \sum_{i=1}^{K} \sum_{j=1}^{K} \rho_{\hat{\tau}_i}^\kappa\big(\delta_{i j}(s,a)\big)$ 
			\State $\triangleright$ Calculate the sample-based regularizer.
			\For{$i=1$ to $N_{r}$}
			\State sample $\{\xi_{i},\xi_{j}\} \sim \mathscr{P}(\mathcal{X})$
			\State sample $\{a_i,a_j\}$ from random policy
			\State $x_i = \pi(s', \xi_{i};\theta_\pi)$,$x_j =\pi(s', \xi_{j};\theta_\pi)$
			\EndFor
			\State calculate $d_e^m(\theta_\pi)$ according to \eqref{equ:mmd-def} or \eqref{eq: P-wass samplebased}
			\State $\triangleright$ Calculate the utility value function.
			\begin{scriptsize}
				\State $V^m_{\psi}(s) = \sum_{i=1}^{K}(\tau_{i+1}-\tau_i)z\big(s,\pi(s,\xi;\theta_{\pi}),\hat{\tau}_i;\theta_z\big)$
			\end{scriptsize}
			\EndFor
			\State $\triangleright$ Construct loss functions.
			\State $\mathcal{L}(\theta_z) = \frac{1}{M} \sum_{m=1}^{M}\mathcal{L}^m(\theta_z)$
			\State $\mathcal{L}(\theta_\pi) = \frac{1}{M} \sum_{m=1}^{M}(-V^m_{\psi}(s) + \alpha d_e^m(\theta_\pi))$ 
			\State update $\theta_z,\ \theta_\pi$ according their losses
			\State $\hat{\theta}_z = \beta\hat{\theta}_z + (1-\beta)\theta_z$
			\EndIf
			\EndFor
			\State \textbf{return} policy $\pi$ 
		\end{algorithmic}
	\end{algorithm}
\end{small}

\subsubsection{Encourager with sample-based metric}
Previous study has provided evidence supporting the enhancement of exploration by incorporating diverse behaviors into policies \cite{haarnoja2017reinforcement}. 
Building upon this idea, the encourager is constructed with a sample-based metric, e.g. MMD or p-Wasserstein distance, between the agent's policy and a reference policy.

\noindent \textbf{Maximum mean discrepancy}. Let $\mathcal{F}$ be a unit ball in a \emph{reproducing kernel Hilbert space (RKHS)} $\mathcal{H}$ defined on a compact metric space $\mathcal{X}$.  Then the MMD between two distributions $\nu$ and $q$ is 
\begin{equation}\label{equ:mmd-def}
	MMD(\nu || q) := \sup\nolimits_{f \in \mathcal{F}} \big (\mathbb{E}_{x \sim \nu}[f(x)] - 
	\mathbb{E}_{y \sim q}[f(y)] \big ).
\end{equation}

Note that MMD has an approximation which solely requires  samples from the distributions and does not demand their density functions explicitly.
That is, given $m$-samples $(x_1,...,x_m)$ from $\nu$ and $n$-samples $(y_1,...y_n)$ from $q$, MMD between $\nu$ and $q$ can be approximated by $D_m(\nu || q) =$
\begin{small}

	\begin{equation}\label{equ:mmd-estimator}
		\begin{aligned}
			\Big[
			\frac{1}{m^2} \scalebox{0.9}{$\displaystyle\sum^{m}_{i, j=1}$} k(x_i, x_j)
			+ \frac{1}{n^2} \scalebox{0.9}{$\displaystyle\sum^{n}_{i, j=1}$} k(y_i, y_j)
			- \frac{2}{mn} \scalebox{0.9}{$\displaystyle\sum^{m, n}_{i, j=1}$} k(x_i, y_j)
			\Big]^{\frac{1}{2}}.
		\end{aligned}
	\end{equation}
\end{small}

\noindent \textbf{P-Wasserstein distance}. The p-Wasserstein distance is formally defined as
\begin{equation}\label{entropy-regularized OT}
	W_{p}(\nu||q)=
	\big(
	\inf_{J \in \mathcal{J} (\nu,q)}
	\int_{\mathcal{X}\times\mathcal{X}}\left\lVert x-y\right\rVert^p \mathrm{d}J(x,y)
	\big)^{\frac{1}{p}},
\end{equation}
where $p \geq 1$, $\mathcal{J}(\nu, q)$ denote all joint distributions $J$ for $(X, Y)$ that have marginals $\nu$ and $q$. In other words, $T_{X_{\sharp}} J=\nu$ and $T_{Y_{\sharp}} J=q$ where $T_X(x, y)=x$ and $T_Y(x, y)=y$. Figure 4
Similarly to MMD, we employ an efficient sample-based approximation of $W_{p}$ in the encourager. Given $(x_1,...,x_m)$ from $\nu$ and $n$-samples $(y_1,...y_n)$ from $q$, the p-Wasserstein distance can be approximated by $D_w(\nu||q) = $
\begin{small}
	\begin{equation} \label{eq: P-wass samplebased}
		\begin{aligned}
			\inf_{{\bf \lambda}} &  \Big\{ \big( \scalebox{0.9}{$\displaystyle\sum^{n}_{i=1}$} \scalebox{0.9}{$\displaystyle\sum^{m}_{j=1}$} \lambda_{i,j}|x_i -y_j|^p \big)^{\frac{1}{p}} + \epsilon \scalebox{0.9}{$\displaystyle\sum^{n}_{i=1}$} \scalebox{0.9}{$\displaystyle\sum^{m}_{j=1}$} \lambda_{i,j} \log (mn\lambda_{i,j}) \Big\}  \\
			& s.t. \scalebox{0.9}{$\displaystyle\sum^{n}_{i=1}$}\lambda_{i,j} = 1/m  \quad \forall j, \quad  \scalebox{0.9}{$\displaystyle\sum^{m}_{i=1}$}\lambda_{i,j} = 1/n \quad \forall i .
		\end{aligned}
	\end{equation}
\end{small}



PACER leverages the uniform policy $u(\cdot|s)$ on the action space $\mathcal{A}$ as the reference policy. In fact,
uniform policy is widely utilized in the RL literature to facilitate exploration of the environment \cite{burda2019exploration,sukhbaatar2018intrinsic}.
Denote the encourager by $D_e \big(\pi(\cdot|s;\theta_\pi)||u(\cdot|s)\big)$, abbreviated as $d_e(\theta_\pi)$. Thus $d_e(\theta_\pi)$ can be chosen from MMD or the p-Wasserstein distance (or other potential sample-based metrics) between $\pi(\cdot|s;\theta_\pi)$ and $u(\cdot|s)$. 
Roughly speaking, the exploration capability of the policy is inversely proportional to $\mathbb{E}_{s \sim d_{\mu 0}^\pi}D_e \big(\pi(\cdot|s;\theta_\pi)||u(\cdot|s)\big)$. By reducing this discrepancy, the encourager can endow the policy with diverse actions, thus effectively incentivizing exploration. 
In practical, Monte Carlo method can be used to estimate the expectation, and the samples of policy $\pi(\cdot|s;\theta_\pi)$ are generated as described previously.

\subsection{Stochastic utility value policy gradient}
Combining the aforementioned components together,
we obtain the objective for the policy in PACER as
\begin{small}
\begin{equation}\label{eq: ACE frame J}
	J_{\psi}\left(\theta_\pi\right) = \mathbb{E}_{s \sim \mu_0}V_{\psi}^{\pi_{\theta}}(s) - \alpha\mathbb{E}_{s \sim \mathcal{B}} D_e \big(\pi(\cdot|s;\theta_\pi)||u(\cdot|s) \big),
\end{equation}
\end{small}
where $\alpha$ denotes the regularizer's weight, and $\mathcal{B}$ is the distribution of sampling from the replay buffer \footnote{We use $s \sim \mathcal{B}$ to approximate $s \sim d_{\mu 0}^\pi$ for MC in empirical study, which is a common practice in RL literatures \cite{lillicrap2015continuous,haarnoja2018soft}.}.
By maximizing $J_{\psi}\left(\theta_\pi\right)$, the policy pursues a large expected utility while simultaneously maintaining exploration due to the MMD regularizer.
Generally, the optimization process in PACER can be divided into two steps. 
We firstly leverage distributional TD learning to update the  critic network parameters, whose loss is defined as follows.
\begin{equation}\label{eq: final loss IDGA}
	\mathcal{L}(\theta_z) = \mathbb{E}_{s\sim d_{\mu_0}^{\pi}, a \sim \pi(\cdot| s;\theta_\pi)} \scalebox{0.9}{$\displaystyle \sum\nolimits_{i=0}^{N-1} \sum\nolimits_{j=0}^{N'-1} $} \rho_{\hat{\tau}_i}^\kappa\big(\delta_{i j}(s,a)\big),
\end{equation}
where $\rho_{\hat{\tau}_i}^\kappa\big(\delta_{i j}(s,a)\big)$ is defined as equation \eqref{eq: quantile loss}, and it can be efficiently optimized using SGD method.
Secondly, we optimize the parameters in the policy  according to $J_{\psi}\left(\theta_\pi\right)$ by leveraging gradient ascent iteratively.

Note that the optimization for the first part of $J_{\psi}\left(\theta_\pi\right)$, i.e. $\mathbb{E}_{s \sim \mu_0}V_{\psi}^{\pi_{\theta}}(s) = \mathbb{E}_{s \sim \mu_0, a \sim \pi(\cdot| s;\theta_\pi)}Q_{\psi}^{\pi_{\theta}}(s,a)$, is non-oblivious. This is primarily due to the stochastic nature of $\pi_\theta$, which affects both the action selection $a \sim \pi(\cdot|s,\theta_\pi)$ and the function $Q_{\psi}^{\pi_\theta}(s,a)$, whose gradient computation is challenging in general. 
When the density of $\pi_\theta$ is calculable, we can compute its gradient according to the stochastic policy gradient theorem \cite{heess2015learning}.
However, as it is intractable to access the density of a push-forward policy with complex DNNs. Thus, we propose a stochastic utility value policy gradient (SUVPG) theorem that can be approximated only based on the samples of a policy. Details of the proofs are deferred until the Appendix \ref{sec: appendix Theoretical study}.
\begin{theorem}[Stochastic utility value policy gradient] \label{th: manuscript SUVPG}
	For a push-forward policy $\pi(s,\xi;\theta_\pi)$ and a differentiable utility function $\psi(\cdot)$, the policy gradient of the state utility function $\mathbb{E}_{s \sim \mu_0}V_{\psi}^{\pi_{\theta}}(s)$ is given by
	\begin{small}
		\begin{equation}
			\begin{aligned}
				& \nabla_{\theta_\pi} \mathbb{E}_{s \sim \mu_0}V_{\psi}^{\pi_{\theta}}(s) \\
				& = \mathbb{E}_{s \sim d_{\mu 0}^\pi, \xi \sim \mathscr{P}(\mathcal{X})}\big[\nabla_{\theta_\pi} \pi\left(s, \xi ; \theta_\pi\right) \cdot \nabla_a Q_{\psi}^{\pi_{\theta}}\left(s, a \right)\vert_{a = \pi(s, \xi;\theta_\pi)}\big].
			\end{aligned}
		\end{equation}
	\end{small}
\end{theorem}

According to Theorem \ref{th: manuscript SUVPG}, it can be verified that the gradient of $J_{\psi}\left(\theta_\pi\right)$ can be calculated as follows.
\begin{small}
	\begin{equation}
		\begin{aligned}
			& \mathbb{E}_{s \sim d_{\mu 0}^\pi, \xi \sim \mathscr{P}(\mathcal{X})}
			\big[ \nabla_{\theta_\pi} \pi\left(s, \xi ;\theta_\pi\right) \cdot \nabla_a Q_{\psi}^{\pi{\theta}}\left(s, a \right) \big] \\
			& - \alpha \mathbb{E}_{s \sim \mathcal{B}} \nabla_{\theta_\pi} D_e \big(\pi(\cdot|s;\theta_\pi)||u(\cdot|s) \big) ,
		\end{aligned}
	\end{equation}
\end{small}

\noindent whose Monte Carlo approximation can be efficiently calculated with only action samples from $\pi_\theta$ by the push-forward policy $\pi\left(s, \xi ; \theta_\pi\right)$. This implies that PACER, unlike algorithms such as SAC, DSAC and IDAC, does not necessitate assumption about the density function associated with the policy $\pi_\theta$ (e.g., Gaussian). 

Actually, SUVPG can be regarded as the policy gradient obtained under the reparameterization technique \citep{kingma2013auto}, while the widely used REINFORCE gradient \citep{heess2015learning} is based on the log-derivative trick.
This also suggests that the push-forward policy is applicable to a wide range of familiar policy gradient based RL algorithms (See Appendix \ref{sec: appendix Implementation of SUVPG in TD3} for an example). 

\subsection{Discussion}
\textbf{Differences between SUVPG and DPG/SPG:}
The proposed SUVPG is non-trivial as previous PG theory (both DPG and SPG) is not applicable to our setting, which involves stochastic policies with utility functions and operates in a fully push-forward DRL frame. 
SUVPG places emphasis on addressing stochastic policies rather than deterministic policies. In comparison to SPG, SUVPG does not require the use of actions' scores or explicit policy density. This aspect enables SUVPG to handle a broader range of policy representations.  
Note that the main purpose of this paper is to fully exploit the expressiveness of push-forward policy in the DRL literature. The main challenge lies in that the intractable density of push-forward policies prevents the usage of existing policy gradient theorem and entropy-type exploration regularizer. As a result, we establish a novel SUVPG theorem to compute the push-forward policy gradient of the cumulative reward and propose a novel sample-based regularization method to incentive exploration, both solely based on samples, eliminating the demand for explicit policy density.

\textbf{Comparison with reparametrization trick in SAC:}
Although the reparameterization trick can be implemented in SAC, the necessity for a policy with tractable density remains unchanged. SAC's policy update employs the Monte Carlo gradient estimate, represented as follows:
$$
\nabla_\theta \log \pi_\theta\left(a | s\right)+\big(\nabla_{a} \log \pi_\theta(a | s)-\nabla_{a} Q(s, a)\big) \nabla_\theta f_\theta\left(\xi ; s\right),
$$ 
where $a = f_\theta\left(\xi ; s\right)$ is the policy employing the reparameterization trick. However, the computation of $\log \pi_\theta\left(a | s\right)$ still relies on having access to the policy density. In other words, this unbiased gradient estimator extends the policy gradients of DDPG style to any stochastic policy that is tractable.
Considering the aforementioned rationales, we denote the policy utilizing the reparameterization trick yet necessitating tractable density as the constrained push-forward policy. 
Indeed, the ongoing focus of research lies in developing policies possessing both high expressiveness and tractable density, such as those based on normalizing flows. Our work goes further beyond this pursuit by entirely obviating the requirement for tractable density.


\textbf{Push-forward policy vs. normalizing flow policy:}
The normalizing flow policy is based on a sequence of bijective transformations, which convert a latent variable following a standard Gaussian distribution into the posterior distribution. Each bijective transformation incurs a high computational cost, primarily attributed to the calculation of inverse Jacobian \cite{kobyzev2019normalizing}. The essence of normalizing flow policy lies in its ability to solve probability densities by restricting the transformations used to bijective ones. However, due to the limited expressive power of bijective transformations, achieving complex distributions often necessitates the utilization of numerous transformations, thereby further increasing the overall computational complexity. 
On the contrary, the push-forward policy can achieve strong modeling capabilities with only shallow networks, eliminating the need for stacking multiple transformations, which surpasses the normalizing flow policy in terms of computational complexity.
In addition, existing Actor-Critic algorithms utilizing normalizing flow policies, such as LSP \cite{haarnoja2018latent} and SAC-NF \cite{mazoure2020leveraging}, are proposed within the maximum entropy framework.
Even without considering the issue of high computational complexity, the expressiveness of the policies in these algorithms are confined by the framework. In contrast, the ACE framework introduced in this paper does not require explicit knowledge of the posterior distribution. 
Consequently, both the normalizing flow policy and the push-forward policy can potentially demonstrate improved performance within our framework.

\textbf{Flexibility of ACE framework:}
The ACE framework is a sampling-based framework that, in comparison to the maximum entropy framework, offers greater flexibility and achieves competitive or even superior performance for exploring environments.
Regarding the actor, our algorithm is particularly suitable for stochastic policies that can be characterized as generative policies \cite{goodfellow2020generative}. Additionally, some advanced network architectures and update methodologies, e.g. transformer \cite{melo2022transformers} and BBF \cite{schwarzer2023bigger}, can also be employed with the actor.
The critic in our algorithm undertakes a twofold responsibility, modeling the value distribution and evaluating the corresponding policy using utility functions. The two primary approaches for distribution parametrization, namely categorical representation \cite{bellemare2017distributional} and quantile representation \cite{dabney2018distributional}, are both adaptable to our algorithm. 
Moreover, the ACE framework accommodates both reward shaping and risk-measure utilities, as elaborated in the Appendix \ref{app: PACER risk measure} for PACER with risk-measure utility. Utility functions should be differentiable to facilitate gradient back propagation. Notice that When adopting expectation as the utility function, the DRL algorithm degenerates into the conventional RL scenario.
In a broader sense, any metric possessing the following two properties is compatible with the encourager: (\romannumeral1) it can be computed solely based on sampling, and (\romannumeral2) it is easily differentiable. As a result, metrics such as maximum mean discrepancy (MMD), Sinkhorn distance \cite{cuturi2013sinkhorn}, Wasserstein distance and its variants, etc. \cite{cuturi2013sinkhorn,peyre2019computational} could serve as alternative options for the encourager.

\section{Experiments} \label{sec: exp}

A comprehensive set of experiments are conducted to demonstrate the performance of PACER on MuJoCo continuous control benchmarks \cite{todorov2012mujoco} and customized navigation tasks by Safety Gym \cite{ray2019benchmarking}.
(\romannumeral1) We evaluate PACER on six MuJoCo continuous control environments and compared its performance with state-of-the-art algorithms.
(\romannumeral2) To verify the effectiveness of the push-forward policy and the sample-based encourager, ablation study is conducted based on the MuJoCo HumanoidStandup environment.
(\romannumeral3) To assess PACER algorithm's efficacy amidst environmental uncertainty with a non-neutral utility function, we evaluate its performance using exponential functions and CVaR as utilities in a modified Halfcheetah environment.
(\romannumeral4) We evaluate the exploration ability and multimodality modeling capability of the push-forward policy through customized navigation tasks in highly stochastic environments, while also assessing the performance of the fully push-forward framework in such environments.

\subsection{Benchmarks}
\begin{figure}[!tb] 
	\centering
	\subfigure[]{
		\includegraphics[width=0.14\textwidth]{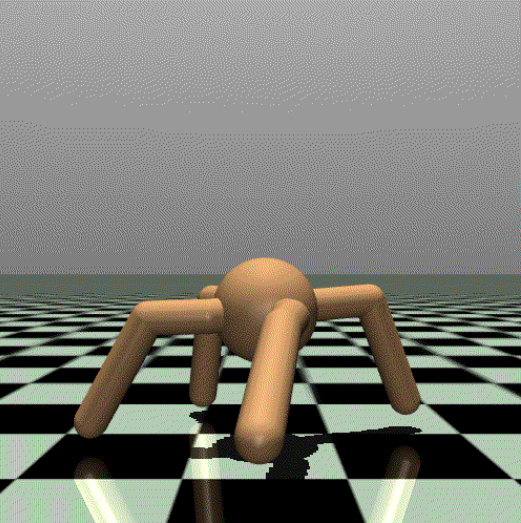} \label{fig:gym-antv2} 
	} 
	\subfigure[]{
		\includegraphics[width=0.14\textwidth]{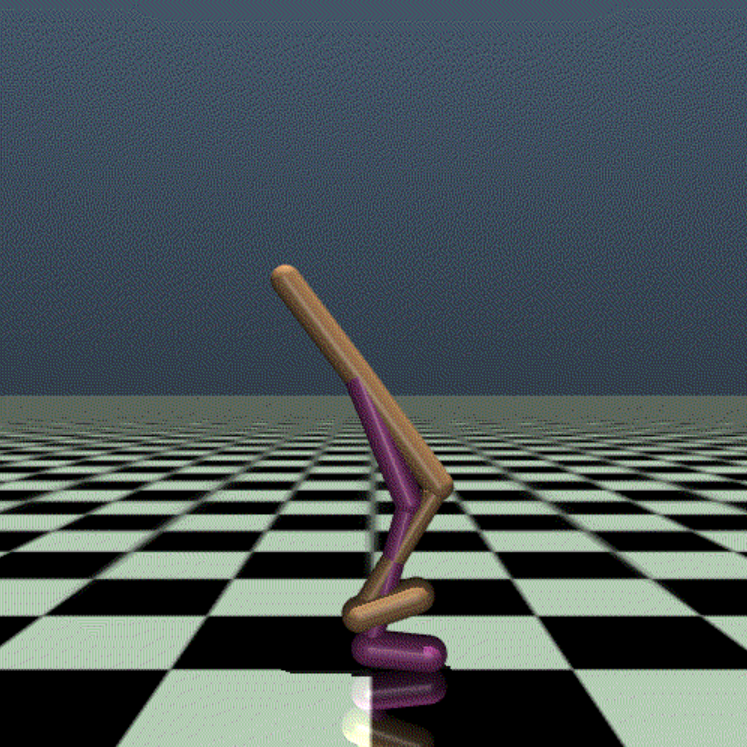} \label{fig:gym-walker}
	}
	\subfigure[]{
		\includegraphics[width=0.14\textwidth]{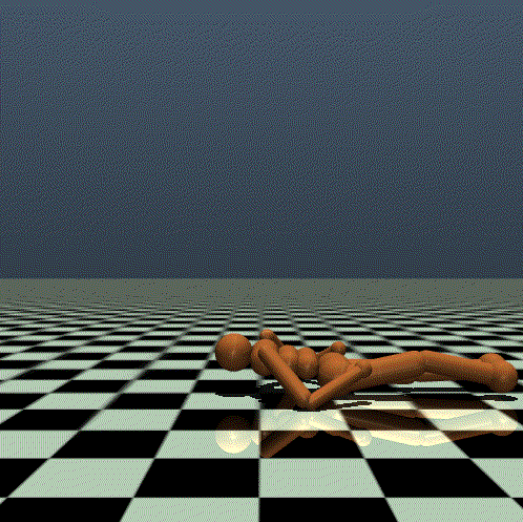} \label{fig:gym-humanstandup} 
	}   
	
	\subfigure[]{
		\includegraphics[width=0.14\textwidth]{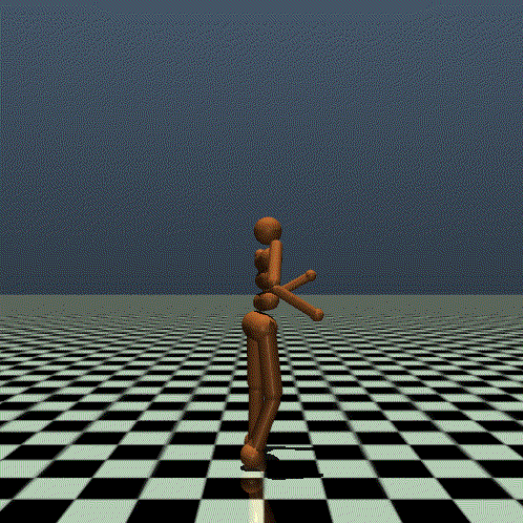} \label{fig:gym-humanoidv2IGDAC} 
	}    
	\subfigure[]{
		\includegraphics[width=0.14\textwidth]{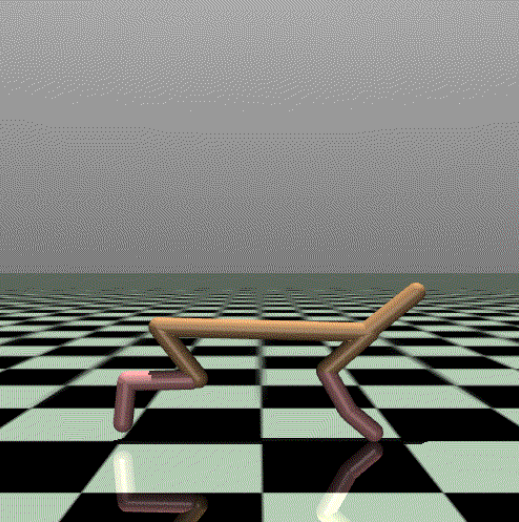} \label{fig:gym-HalfCheetah}
	}
	\subfigure[]{
		\includegraphics[width=0.14\textwidth]{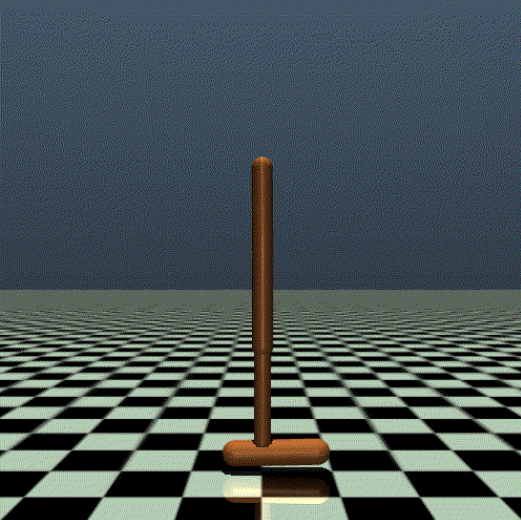} \label{fig:gym-Hopper} 
	}       
	\caption{Six MuJoCo continuous control environments. (a) Ant, (b) Walker2d, (c) HumanoidStandup, (d) Humanoid, (e) HalfCheetah, (f) Hopper.} \label{fig: gym env}
\end{figure}

MuJoCo stands for Multi-Joint dynamics with Contact, which is the most representative benchmark for continuous action space problems. Fig. \ref{fig: gym env} shows the continuous control tasks used for experiment (\romannumeral1), (\romannumeral2) and (\romannumeral3), and Table \ref{table: mujoco space} provides the information for their action spaces and observation shapes. Generally speaking, the complexity of a task is positively correlated with the size of its space $|\mathcal{A}||\mathcal{O}|$.
Details for environment reward can be found in the official document\footnote{https://www.gymlibrary.dev/environments/mujoco/}. 

\begin{table}[!tb]
	\centering
	\caption{Action space ($\mathcal{A}$) and observation space ($\mathcal{O}$) details} \label{table: mujoco space}
	\begin{tabular}{lcc}
		\toprule
		Environment & Action space $\mathcal{A}$ & $\mathcal{O}$ Shape \\
		\midrule
		Ant & Box(-1.0,1.0,(8,), float32) & (27,) \\
		Walker2d & Box(-1.0,1.0,(6,), float32) & (17,) \\
		H-Standup & Box(-0.4, 0.4, (17,), float32) & (376,)\\
		Humanoid & Box(-0.4,0.4,(17,), float32) & (376,) \\
		Halfcheetah & Box(-1.0,1.0,(6,), float32) & (17,) \\
		Hopper & Box(-1.0,1.0,(3,), float32) & (11,) \\
		\bottomrule
	\end{tabular}
\end{table}

\begin{figure}[!tb] 
	\centering
	\subfigure{
		\includegraphics[width=0.22\textwidth,height=0.22\textwidth]{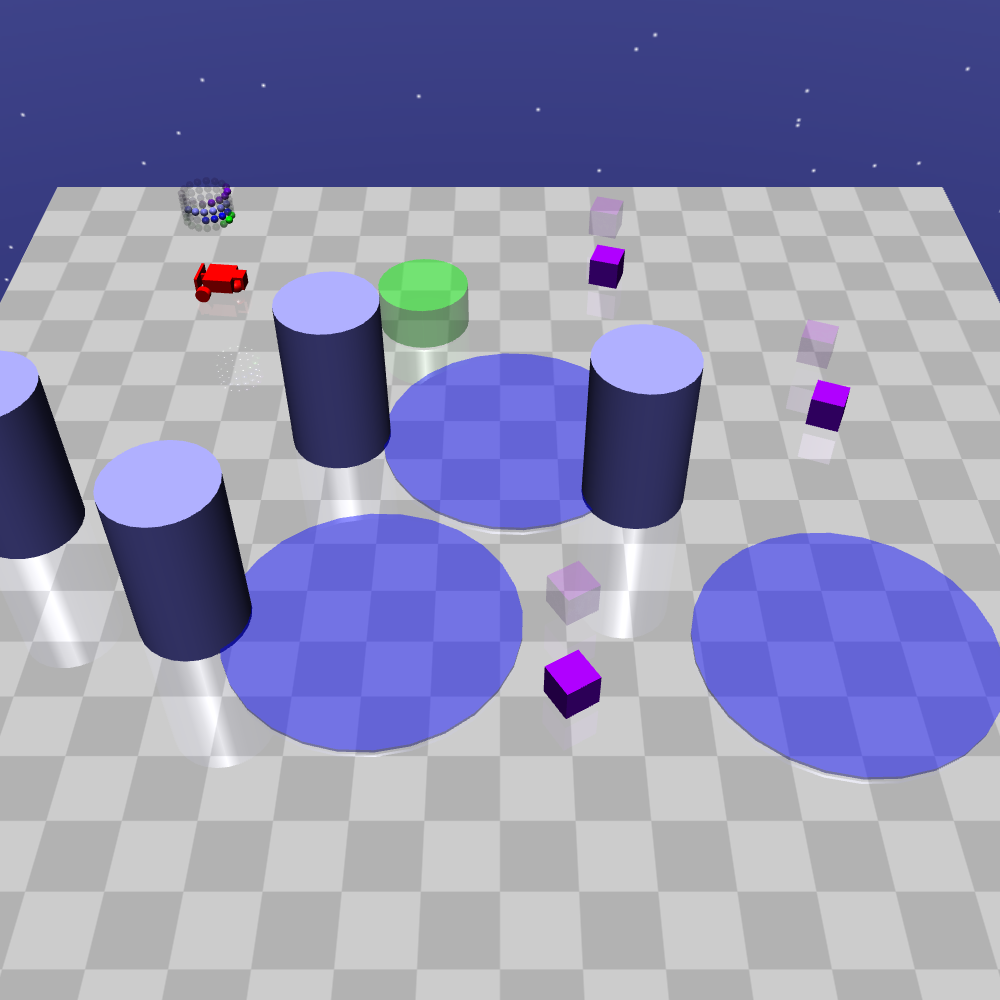} \label{fig:customenv1} 
	} 
	\subfigure{
		\includegraphics[width=0.22\textwidth,height=0.22\textwidth]{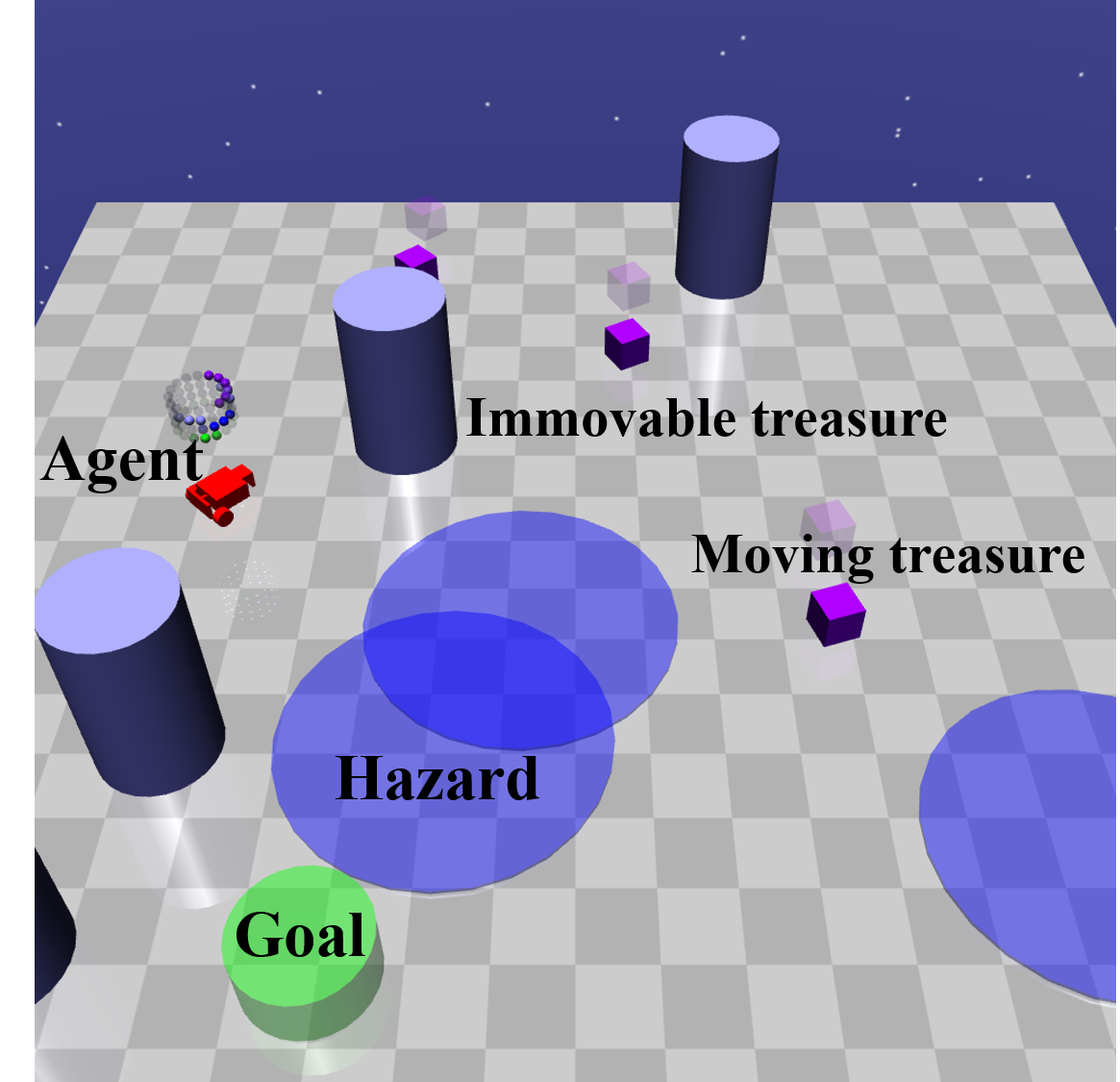} \label{fig:customenv2}
	} 
	\caption{Visualization of the navigation task.} \label{fig: custom env1}
\end{figure}

In experiment (\romannumeral4), we evaluate our algorithm in customized navigation tasks with different complexity levels from the Safety Gym benchmark suite \cite{ray2019benchmarking}. The agent is required to navigate to a target zone on a 2-D plane, avoiding hazards while striving to collect treasures along the way, see Fig. \ref{fig: custom env1}. Our customized tasks introduce novel elements wherein penalties for encountering hazards and rewards for obtaining treasures manifest in probabilistic forms, rather than being deterministic events. Roughly speaking, $R(s,a)$ corresponds to 
$$
\begin{aligned}
	&\text{rewards when moving to goal} - \mathscr{P}(R_{\text{hazard}}) \\
	&+ \mathscr{P}(R_{\text{treasure}})  + \text{reward when reaching goal}, 
\end{aligned}
$$
where $\mathscr{P}(R_{\text{hazard}})$ and $\mathscr{P}(R_{\text{treasure}})$ are random variable reward for encountering hazards and obtaining treasures, following pre-defined distributions (refer to Appendix \ref{app: Customized environments details} for environment details).
This design results in complex return distributions. Crucially, the broad distribution of treasures in the environment allows the agent to easily acquire small rewards during training. Thereby, without a policy that incorporates diversity to explore, the agent risks falling into local optima, neglecting the substantial reward associated with achieving the ultimate goal. The complexity of these tasks is further compounded by the nature of the observation space, wherein the agent does not directly observe its location, but instead relies on a lidar sensor that provides information about the distances to other objects in the environment.

\subsection{Baselines}
Our baselines include: Implicit Distributional Actor Critic (IDAC) \cite{yue2020implicit}, Distributional Soft Actor Critic (DSAC) \cite{ma2020dsac}; As well as expectation-based Actor-Critic algorithms, SAC \cite{haarnoja2018soft}, LSP \cite{haarnoja2018latent}, SQL \cite{haarnoja2017reinforcement}, TD3 \cite{fujimoto2018addressing}, and DDPG \cite{lillicrap2015continuous}. 
\begin{enumerate}
	\item IDAC: It is the SOTA DAC algorithm that learns a semi-implicit stochastic policy modeled by a mixture of Gaussians, with a distributional critic modeled by IQN.
	\item DSAC: It is a distributional version of SAC, with an entropy regularized distributional critic modeled by IQN.
	\item SAC: It is an off-policy AC algorithm that learns a stochastic policy with parametric distribution modeled by push-forward operator, with an entropy regularized Q target.
	\item LSP: It is an off-policy AC algorithm that learns a hierarchical stochastic policy modeled by normalizing flows, with an entropy regularized Q target.
	\item SQL: It is an off-policy AC algorithm that learns a stochastic policy using Stein Variational Gradient Descent, with a soft target for the Q-value function that follows the maximum entropy principle.
	\item TD3: It is an off-policy AC algorithm that utilizes the clipped double Q-learning and delayed policy updates to alleviate the overestimation errors in DDPG.
	\item DDPG: It is an off-policy AC algorithm where the actor uses a deterministic policy whose gradient is conditional on the learned critic.
\end{enumerate}
\begin{table*}[tbh]
	\centering
	\caption{Comparison of max average returns $\pm$ 1 std over 7 different random seeds.} \label{tab: average return}
	\begin{tabular}{lcccccc}
		\toprule
		& Ant & Walker2d & HumanoidStandup & Humanoid & HalfCheetah & Hopper\\ 
		\midrule
		PACER$_M$ & 6386 $\pm$ 378 & \textbf{5500 $\pm$ 123} & \textbf{198856 $\pm$ 38610} & 6094 $\pm$ 596 & \textbf{12062 $\pm$ 1403} & \textbf{3452 $\pm$ 152} \\ 
		PACER$_W$ &  \textbf{6501$\pm$172}  &  5499$\pm$117  &  194292$\pm$33118  &  \textbf{7668$\pm$920}  &  7215$\pm$602  &  3328$\pm$120 \\ 
		IDAC & 5160 $\pm$ 920 & 5182 $\pm$ 471 & 187602 $\pm$ 29975 & 5860 $\pm$ 31 & 8985 $\pm$ 1673 & 2958 $\pm$ 608 \\ 
		DSAC & 1981 $\pm$ 1478 & 2889 $\pm$ 1639 & 168293 $\pm$ 23916 & 2350 $\pm$ 1496 & 5633 $\pm$ 815 & 3309 $\pm$ 160\\ 
		SAC  & 5432 $\pm$ 357 & 4654 $\pm$ 310 & 147349 $\pm$ 11908 & 5128 $\pm$ 78 & 11356 $\pm$ 616 & 3219 $\pm$ 292\\ 
		LSP  & 3936 $\pm$ 427 & 5029 $\pm$ 466 & 141823 $\pm$ 17658 & 5367 $\pm$ 148 & 8200 $\pm$ 567 & 3358 $\pm$ 376\\ 
		SQL  & 1483 $\pm$ 280 & 3032 $\pm$ 329 & 167320 $\pm$ 28969 & 1663 $\pm$ 701 & 5461 $\pm$ 649 & 2803 $\pm$ 286\\ 
		TD3  & 4998 $\pm$ 583 & 4677 $\pm$ 587 & 79566 $\pm$ 25874 & 5023 $\pm$ 382 & 9305 $\pm$ 1331 & 2696 $\pm$ 1351\\ 
		DDPG & 1015 $\pm$ 800 & 2637 $\pm$ 1032 & 69131 $\pm$ 28895 & 130 $\pm$ 101 & 6525 $\pm$ 2336 & 1669 $\pm$ 930\\ 
		\bottomrule
	\end{tabular}
\end{table*}

\begin{figure*}[tbh] 
	\centering
	\subfigure{
		\includegraphics[width=0.30\textwidth]{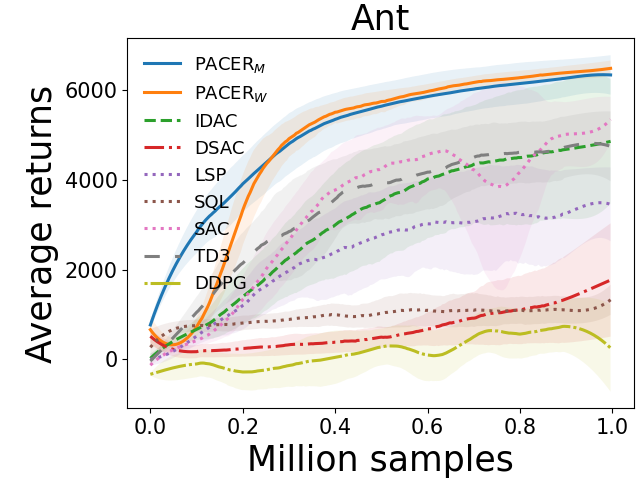} \label{fig:antv2 } 
	} 
	\subfigure{
		\includegraphics[width=0.30\textwidth]{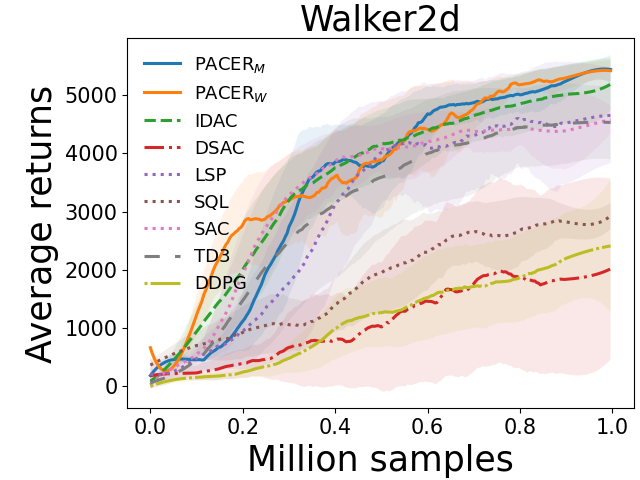} \label{fig:walker }
	}
	\subfigure{
		\includegraphics[width=0.30\textwidth]{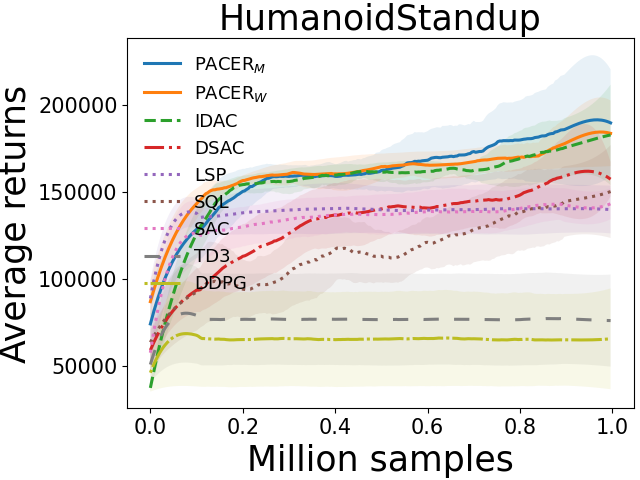} \label{fig:humanstandup } 
	}   
	
	\subfigure{
		\includegraphics[width=0.30\textwidth]{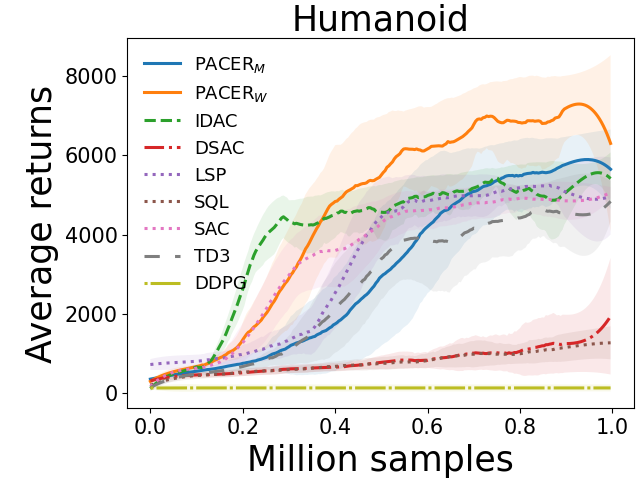} \label{fig:humanoidv2IGDAC } 
	}    
	\subfigure{
		\includegraphics[width=0.30\textwidth]{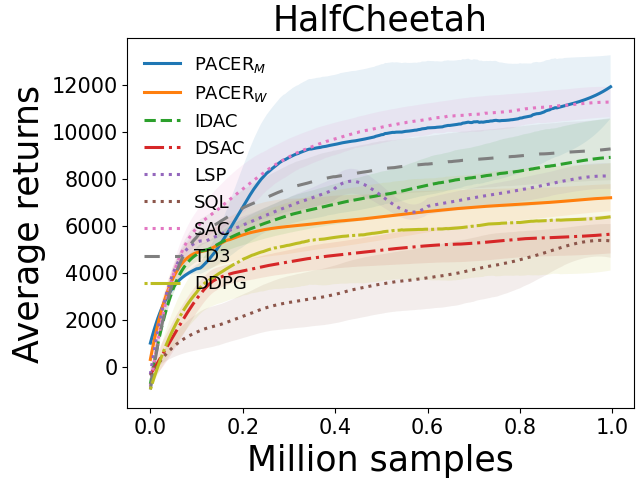} \label{fig:HalfCheetah }
	}
	\subfigure{
		\includegraphics[width=0.30\textwidth]{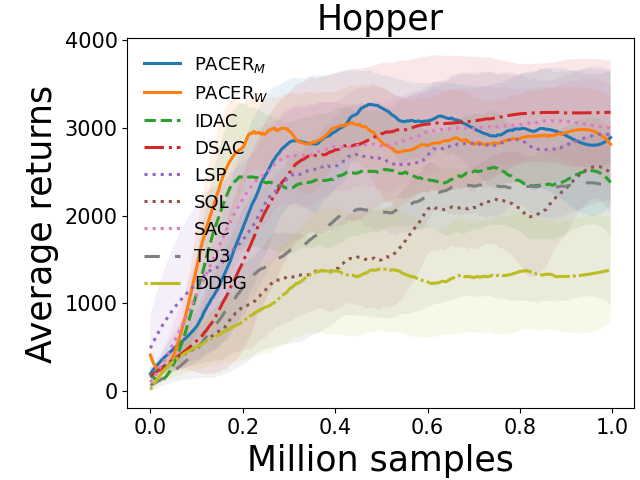} \label{fig:Hopper } 
	}       
	\caption{Learning curves for PACER and baselines with $\pm$ 1 std shaded on MuJoCo continuous control tasks.} \label{fig: main training curves}
\end{figure*}

To implement the baselines, we use codes from websites provided in the original papers for IDAC\footnote{https://github.com/zhougroup/IDAC} \cite{yue2020implicit}, DSAC\footnote{https://github.com/xtma/dsac} \cite{ma2020dsac} and SQL\footnote{https://github.com/haarnoja/softqlearning} \cite{haarnoja2017reinforcement}, and we modify the code provided by SpinningUp\footnote{https://github.com/openai/spinningup} \cite{SpinningUp2018} for SAC, DDPG and TD3. For LSP, in the absence of an official implementation, we reproduced the algorithm following the description from the original paper \cite{haarnoja2018latent}.
We incorporate twin delayed networks and target networks into all algorithms, and neutral utility functions (identity maps) are adopted across all DRL algorithms for fair comparison, unless otherwise stated.
We fix the batch size and total environment interactions for all the algorithms, and other tunable hyper-parameters are either set to their best values according their original papers (if provided) or tuned with grid search on proper intervals. 

We implemented two versions of PACER algorithms using MMD and entropy regularized p-Wasserstein distance as the encourager, and denote them by PACER$_M$ and PACER$_W$ respectively. The energy kernel is applied to calculate the MMD. The default version for PACER is PACER$_M$, as the computational complexity for MMD is lower than Wasserstein distance. Network structures and other key hyper-parameters are detailed in Appendix \ref{sec: appendix Experimental details}. 
All experiments are conducted on Nvidia GeForce RTX 2080 Ti graphics cards, aiming to eliminate the performance variations caused by discrepancies in computing power.
We train 10 different runs of each algorithm with 7 different random seeds. The evaluations are performed every 50 steps by calculating their averaged returns. The total environment interactions are set to 1 millions, and the networks' parameters are updated every 50 new samples.

\subsection{Experimental results}

\subsubsection*{(\romannumeral1) Performance compared to baselines}
The numerical results of maximum average returns across all algorithms are given in Table \ref{tab: average return}, and their learning curves are shown in Fig. \ref{fig: main training curves}. The results show that PACER with both MMD and Wasserstein distance achieved exceptionally high scores. The highest scores for all benchmarks have originated from PACER$_M$ and PACER$_W$, and remarkably, PACER$_M$ outperforms all other baselines. Particularly, PACER$_M$ gains approximately $10\%$ performance improvement compared to the existing SOTA. Additionally, the averaged scores for DRL algorithms (PACER$_M$, PACER$_W$, IDAC, DSAC) are higher than those for Non-DRL algorithms (SAC, LSP, SQL, TD3, DDPG), which further demonstrates the advantage of modeling return distributions.

Regarding the performances on the Hopper task, we analyzed it from the perspective of task complexity. Table \ref{fig: custom env1} shows that space size $|\mathcal{A}||\mathcal{O}|$ of Hopper is smaller than other tasks, implying that Hopper is relatively easier. Moreover, the small action space and observation shape may suggest that the value function is unimodal and concentrated \cite{liu2024distributional}. The experimental results exhibit that all baseline algorithms can performance well on the Hopper task, yet they could not attain performance competitive with PACER on more complex tasks. This provides evidence that previous methods may perform well on relatively easy tasks, whereas the proposed PACER algorithms can excel in more complex tasks.

\subsubsection*{(\romannumeral2) Ablation studies}
We run a comprehensive set of ablation studies on the HumanoidStandup task to demonstrate the effectiveness of push-forward actor and the sample-based encourager. The experiment results are given in Table \ref{tab: ablation table} and Fig. \ref{fig: ablation_Ant-v2}. 
We can see PACER that leverages both components (PACER$_M$ and PACER$_W$) outperforms all other ablated algorithms that missing one or more of those components. Moreover, PACER$_M$/PACER$_W$ outperforms M1P0/W1p0, which reveals the improvement of utilizing push-forward policies over Gaussian policies. Besides, we discover that algorithms using sample-based regularizers for exploration outperform those using just $\epsilon\quad greedy$ or even none exploration strategy, i.e. PACER$_M$/PACER$_W$/M1P0/W1P0 achieves higher scores than M0P1G/M0P1/M0P0G/M0P0. 
The results exhibit the significance and effect of adopting push-forward policies and sample-based regularizers in continuous control tasks. Additionally, the results also reveal that: 
(\romannumeral 1) The sample-based regularizers are also effective in promoting exploration for Gaussian-type policies, as evidenced by the second and third-place rankings achieved by M1P0 and W1P0.
(\romannumeral 2) The absence of these crucial components significantly increases the probability of low performance or even failure. 
These findings offer compelling evidence for the effectiveness and significance of incorporating the push-forward policies and sample-based regularizers within DRL algorithms.

\begin{table}[!tb]
	\centering
	\caption{Ablation results} \label{tab: ablation table}
	\begin{tabular}{lcccc}
		\toprule
		& Policy & Exploration & Frame & Score \\
		\midrule
		PACER$_M$ & push-forward & MMD & ACE & \textbf{198856}
		\\
		PACER$_W$ & push-forward & Wasserstein & ACE & 194292
		\\
		M1P0 &  Gaussian & MMD & ACE & 181513
		\\
		W1P0 &  Gaussian & Wasserstein & ACE & 173910
		\\
		M0P1G &  push-forward & $\epsilon$-greedy & AC & 157777
		\\
		M0P1 &  push-forward & None & AC & 140210
		\\
		M0P0G &  Gaussian & $\epsilon$-greedy & AC & 165686
		\\
		M0P0 &  Gaussian & None & AC & 161008
		\\
		\midrule
		DSAC &  Gaussian & KL & AC & 168293
		\\
		\bottomrule
	\end{tabular}
\end{table}

\begin{figure}[!tb]  
	\centering
	\includegraphics[width=0.35\textwidth]{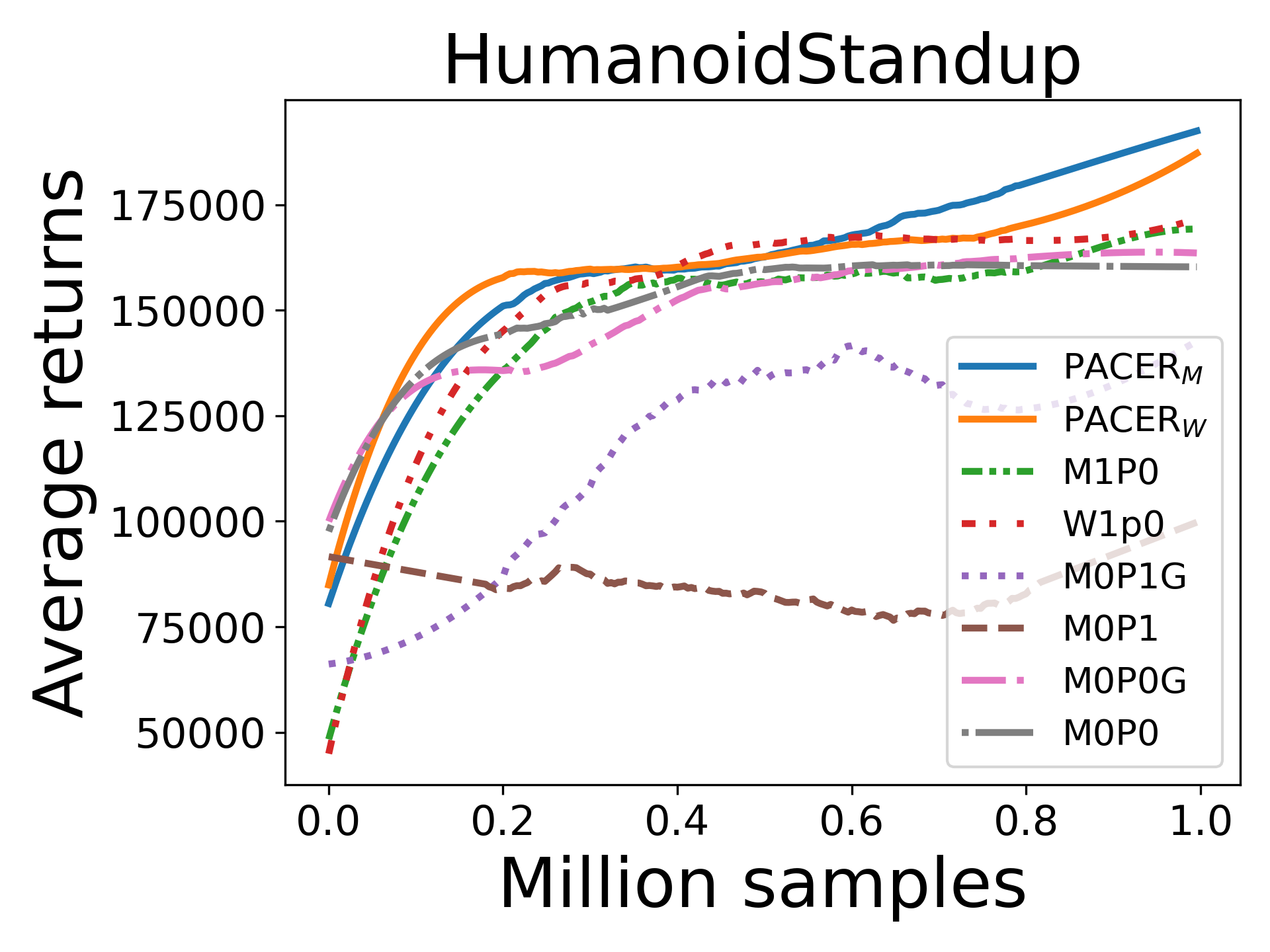} \label{fig: ablation}
	\caption{Learning curves for ablation studies.}
\end{figure}

\subsubsection*{(\romannumeral3) Effectiveness for utility functions} 
To examine the performance of the PACER algorithm in the presence of environmental uncertainty when employing a non-neutral utility function.
We modify the reward function in the HalfCheetah task to have a meaningful assessment of uncertainty and risk following the same idea as \cite{armengol2021risk}. Specifically, $R_t(s, a) = \bar{r}_t(s, a) - 70 \mathbb{I}_{v > 4} \cdot \mathcal{B}_{0.1}$, where $R_t(s, a)$ is the modified reward, $\bar{r}_t(s, a)$ is the original reward, and $v$ is the forward velocity. This modification will penalize high velocities ($v>4$) with a Bernoulli distribution ($\mathcal{B}_{0.1}$), which represents rare but catastrophic events.

\begin{figure}[!tb]  
	\centering
	\subfigure[Exponential utility]{
		\includegraphics[width=0.22\textwidth]{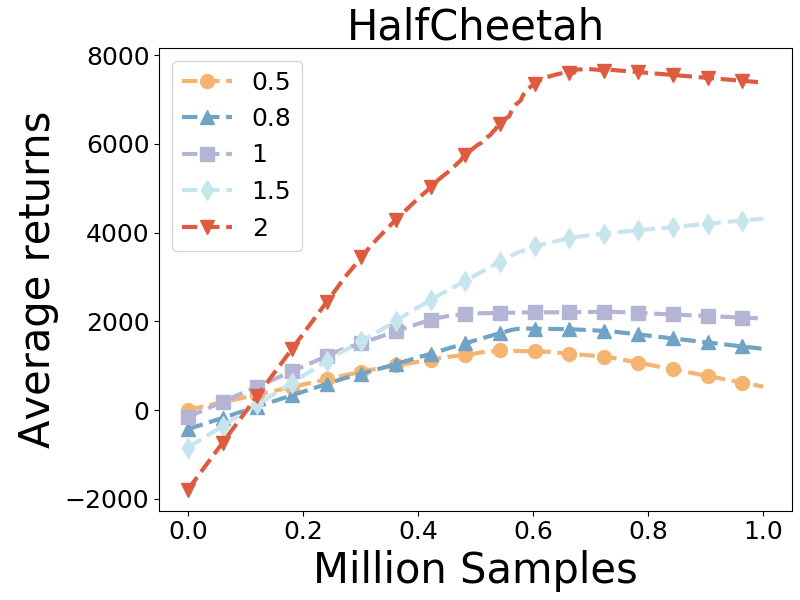} \label{fig: ablation_Ant-v2}
	} 
	\subfigure[CVaR]{
		\includegraphics[width=0.22\textwidth]{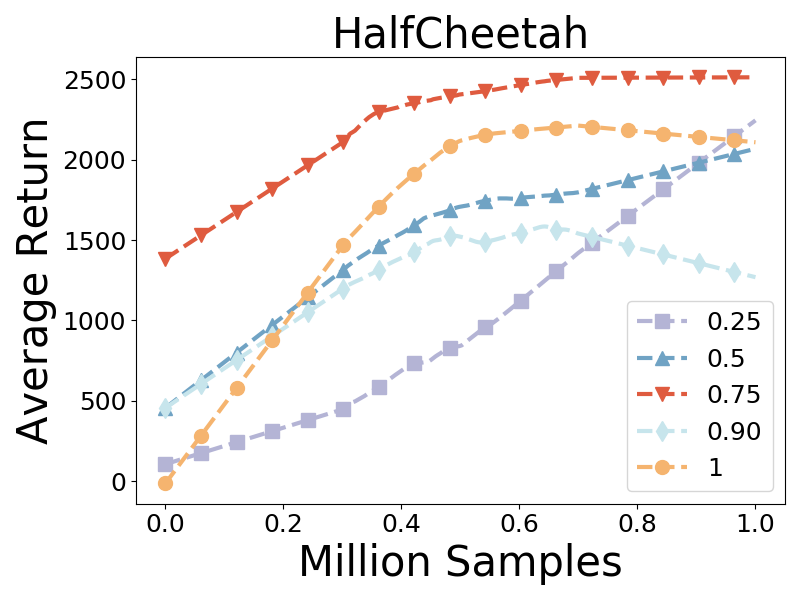} \label{fig: risk}
	} 
	\caption{Learning curves for PACER using exponential utility functions (a) and CVaR utility functions (b).} \label{fig: risk all}
\end{figure}

We evaluate the performance of PACER using exponential utility functions with different powers as examples for reward shaping type utility functions, i.e. $\psi(R) = R^\alpha$, and $Z_{\psi}^{\pi}(s,a) = \big[\displaystyle\sum\nolimits_{t=0}^{\infty}\gamma^tR(s_t,a_t)^{\alpha}\big]$ ($\alpha = 0.5, 0.8, 1, 1.5, 2$).
The learning curves are shown in Fig. \ref{fig: ablation_Ant-v2}, it can be observed that there is a notable increase in the maximum average return as $\alpha$ is augmented.
The results show that the agent seeks to pursue higher returns while concurrently managing the risk, further demonstrating the effectiveness for leveraging utility functions in PACER to generate risk-sensitive policies. 

For risk-measure type utility functions, CVaR is frequently applied in DRL algorithms to regulate the actor's orientation toward either a reward-seeking or a risk-averse policy \cite{ma2020dsac}. It determines the upper bound for the integration of $V_\psi(s)$ under our setting, i.e. assume $\text{CVaR} \in [0,1]$, $V_\psi^\pi(s) = \mathbb{E}_{\xi \sim \mathscr{P}(\mathcal{X})} \int_{0}^{\text{CVaR}} F_{Z(s,\pi(s,\xi;\theta_\pi))}^{-1}(\tau)d\tau$.
While CVaR typically operates directly on $Z(s,a)$ rather than on $R(s,a)$, it can be viewed as an extension of the reward shaping type utility in a loose sense. Furthermore, a considerable body of work in the field of risk-sensitive DRL algorithms regards this criterion as the most significant evaluation metric, with ample empirical evidence supporting its compatibility with DRL settings \cite{dabney2018implicit, ma2020dsac, yang2023safety}, i.e. compatible with our algorithm. Therefore, we evaluate the performance of PACER using different levels of CVaR (${0.25,0.5,0.75,0.90,1}$) as utility functions on the modified HalfCheetah environment.
The results are shown in Fig. \ref{fig: risk}. It is evident that the policy with a 0.75-CVaR outperforms the risk-neutral policy (1-CVaR), since the actor employing the 0.75-CVaR policy demonstrates risk aversion towards the infrequent yet catastrophic event that robot breakdowns. 
The result shows that PACER with proper utility functions has the ability to obtain risk-sensitive policies. 
Additionally, we observe that the overall performance of CVaR is lower compared to exponential utility functions with $\alpha = 1.5$ and $2$. This discrepancy may stem from CVaR causing overly conservative strategies, thereby neglecting actions that could yield higher returns with acceptable risk.

\begin{figure*}[!htb] 
	\centering
	\subfigure{
		\includegraphics[width=0.23\textwidth]{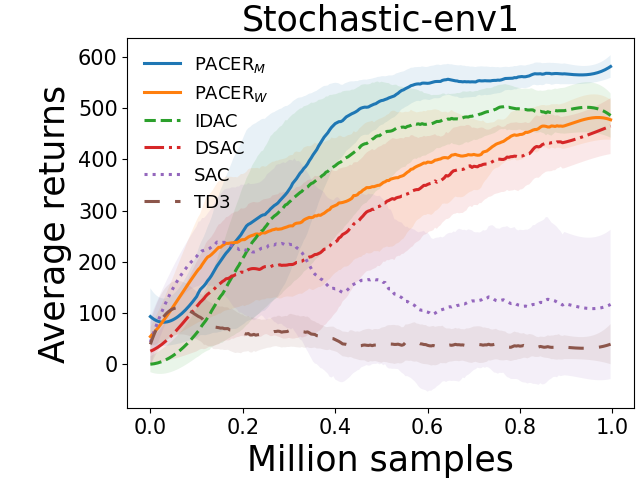} 
	}
	\subfigure{
		\includegraphics[width=0.23\textwidth]{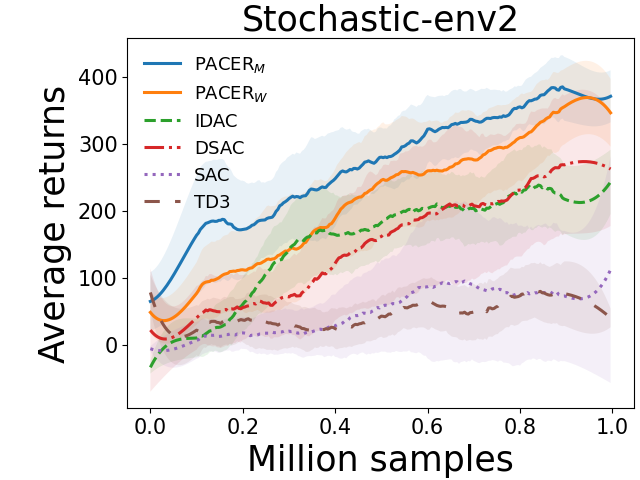} 
	}    
	\subfigure{
		\includegraphics[width=0.23\textwidth]{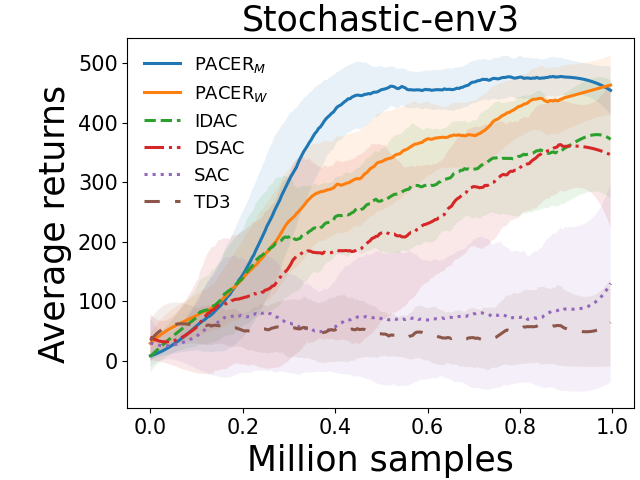} 
	}
	\subfigure{
		\includegraphics[width=0.23\textwidth]{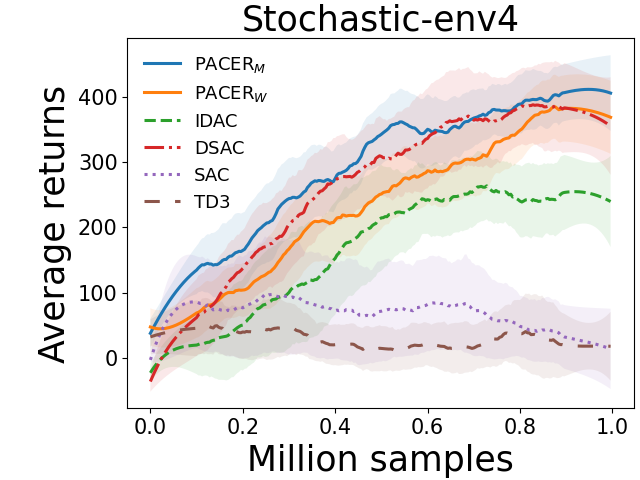} 
	}
	\caption{Learning curves for PACER and baselines with $\pm$ 1 std shaded on customized navigation tasks.} \label{fig: train curve full customized_envs}
\end{figure*}

\begin{figure}[htb] 
	\centering
	\subfigure[Early phase]{
		\includegraphics[width=0.143\textwidth,height=0.11\textwidth]{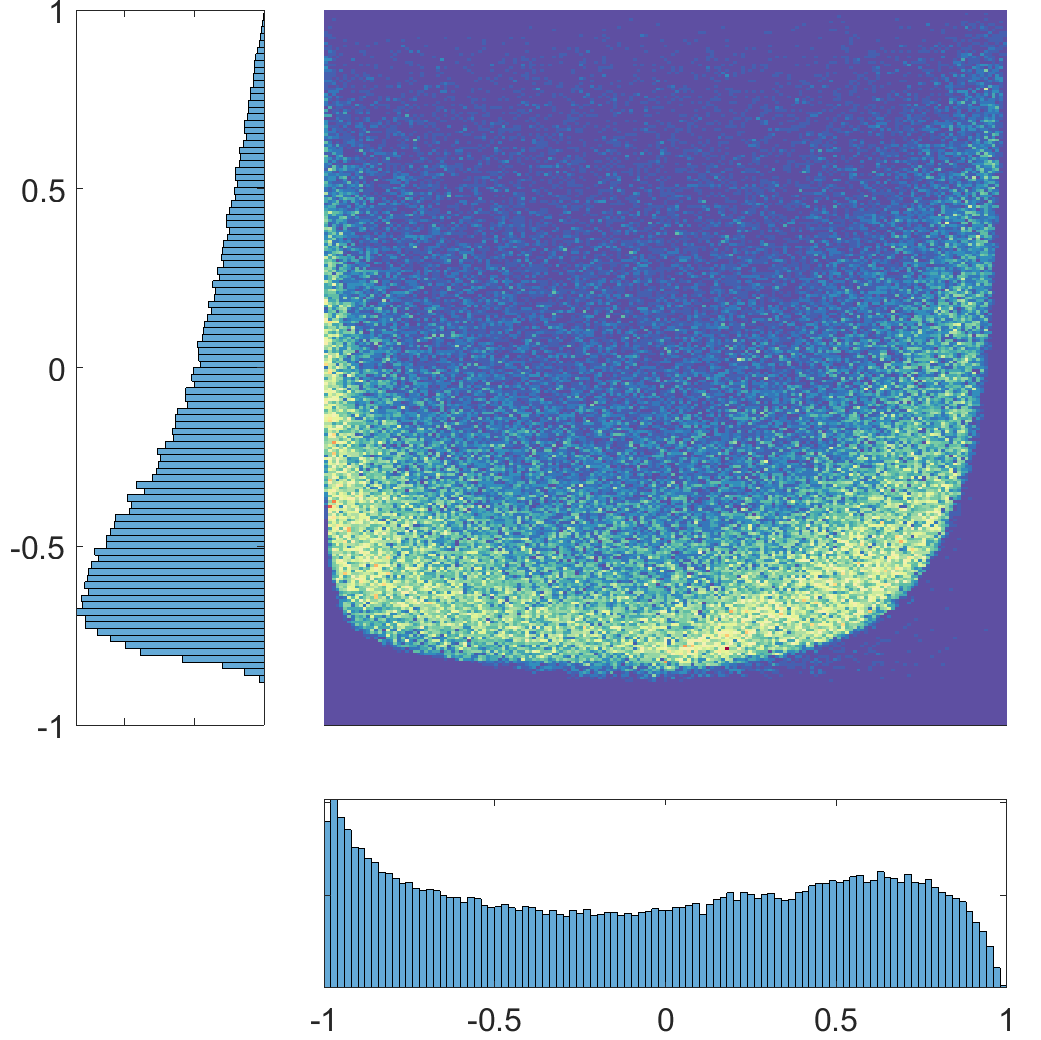} 
	}
	\subfigure[Middle phase]{
		\includegraphics[width=0.143\textwidth,height=0.11\textwidth]{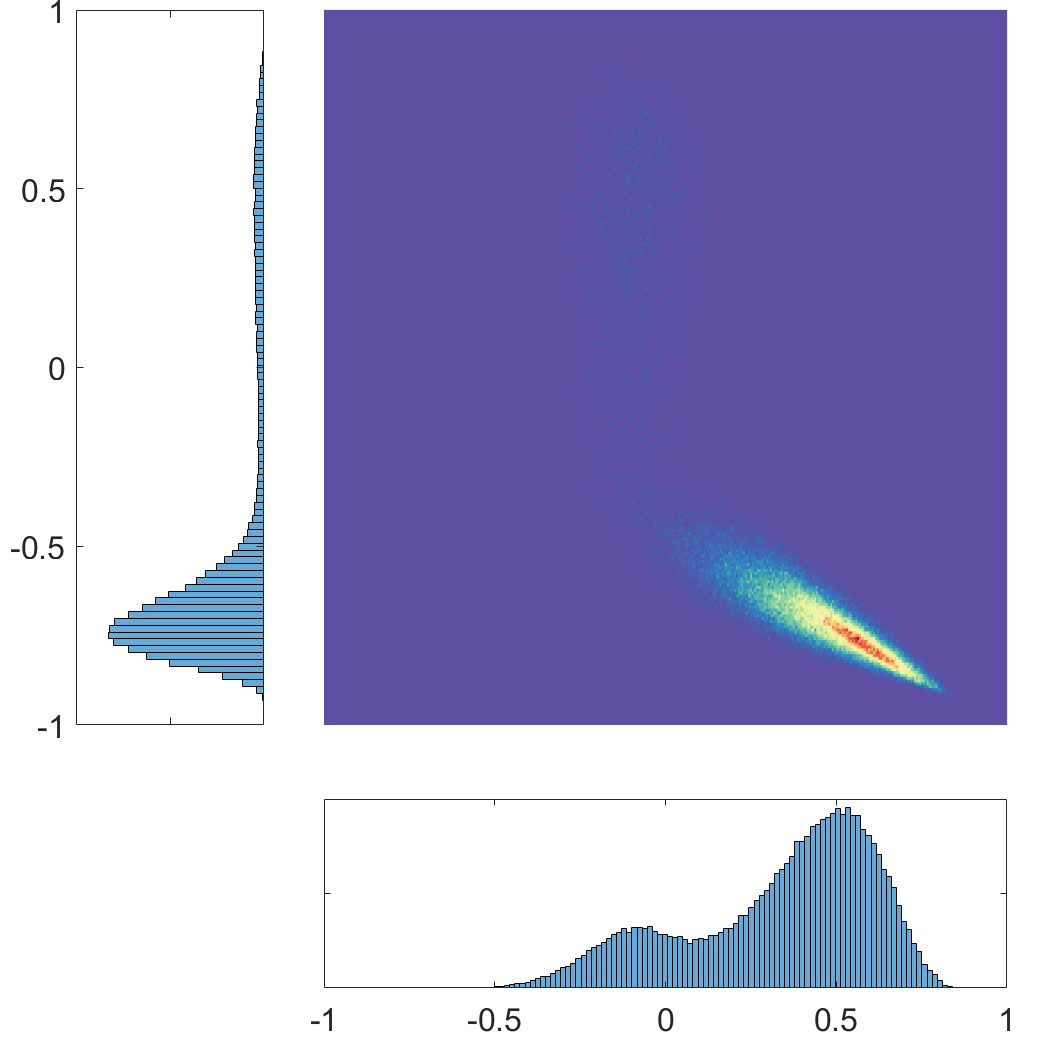} 
	}    
	\subfigure[Final phase]{
		\includegraphics[width=0.143\textwidth,height=0.11\textwidth]{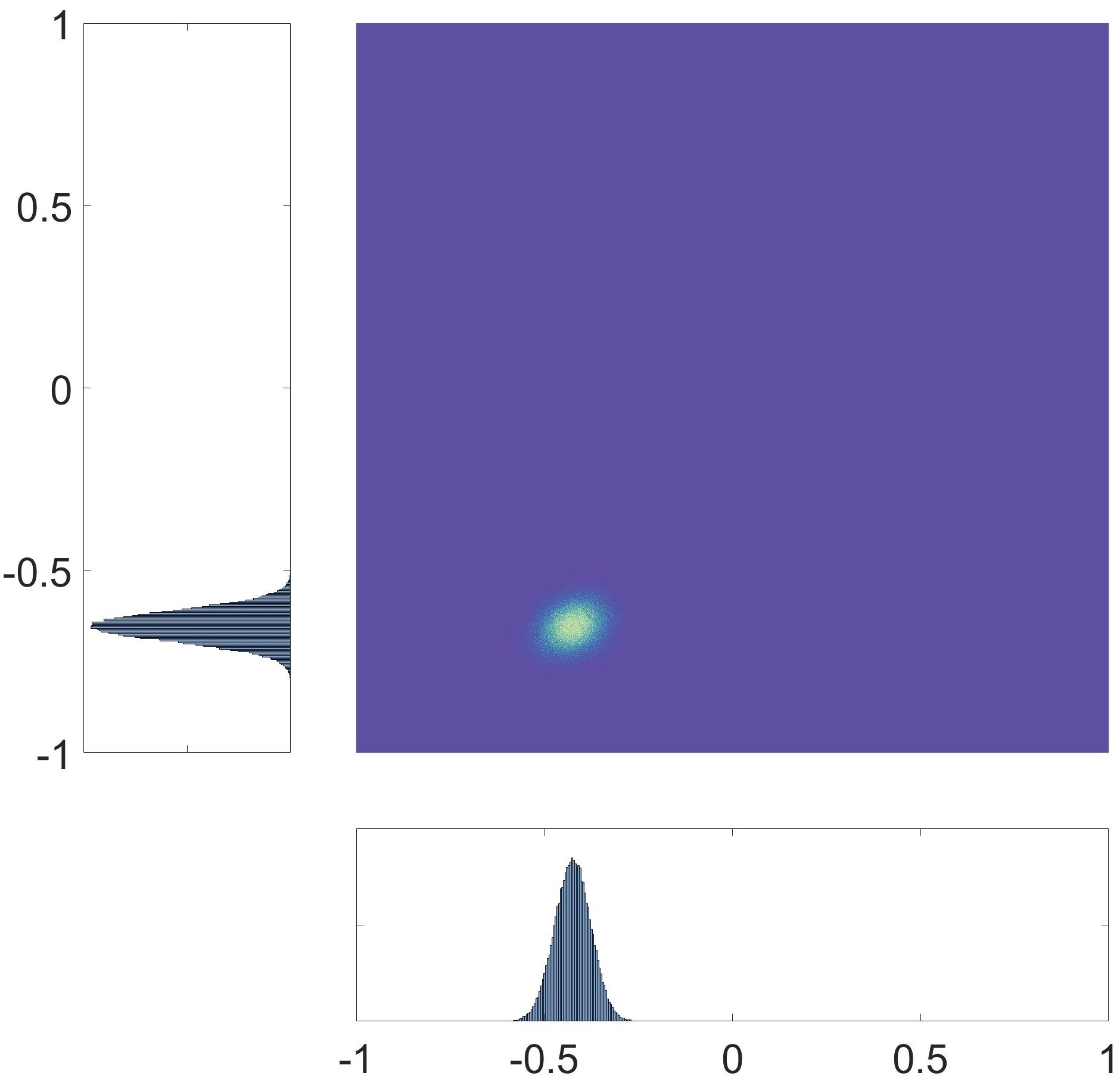} 
	}    
	\\
	\subfigure[Early phase]{
		\includegraphics[width=0.143\textwidth,height=0.11\textwidth]{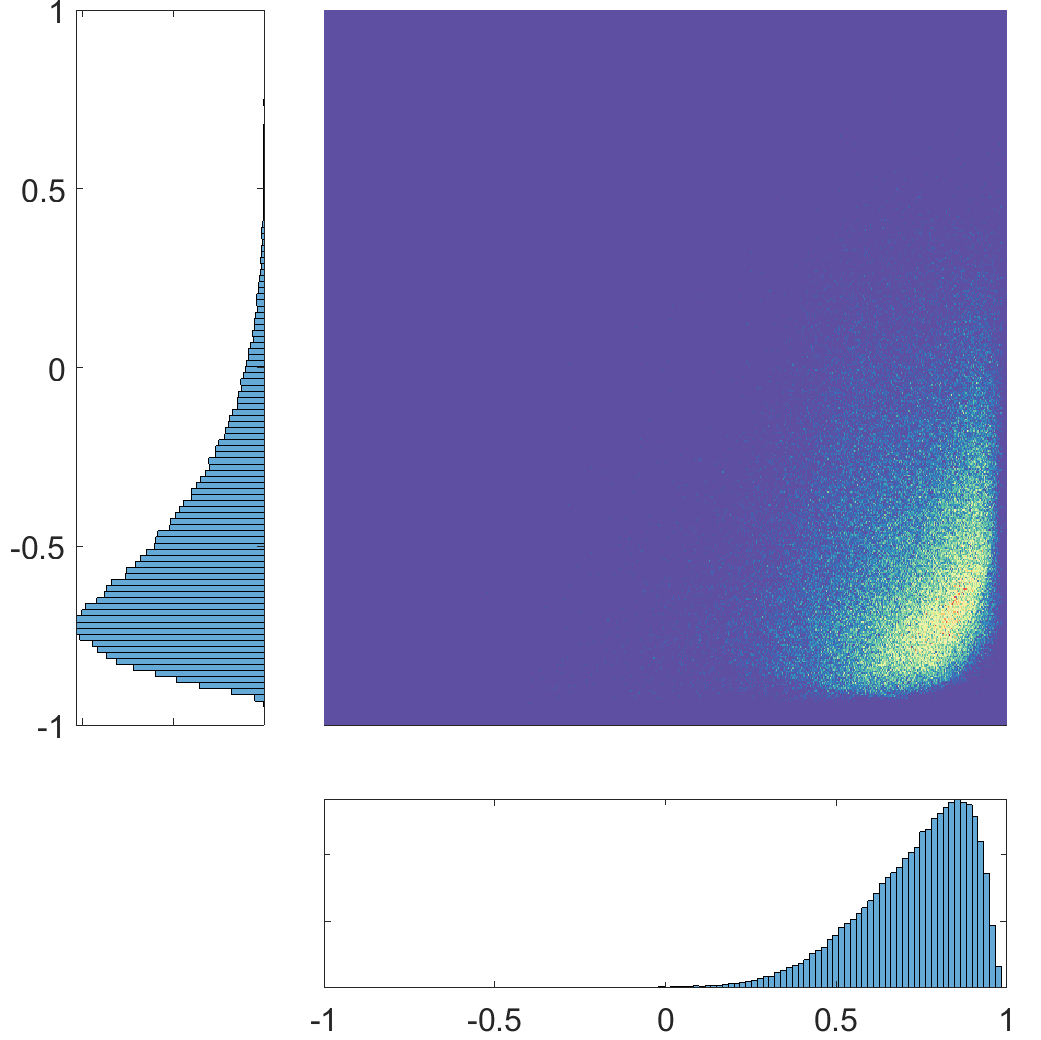} 
	}
	\subfigure[Middle phase]{
		\includegraphics[width=0.143\textwidth,height=0.11\textwidth]{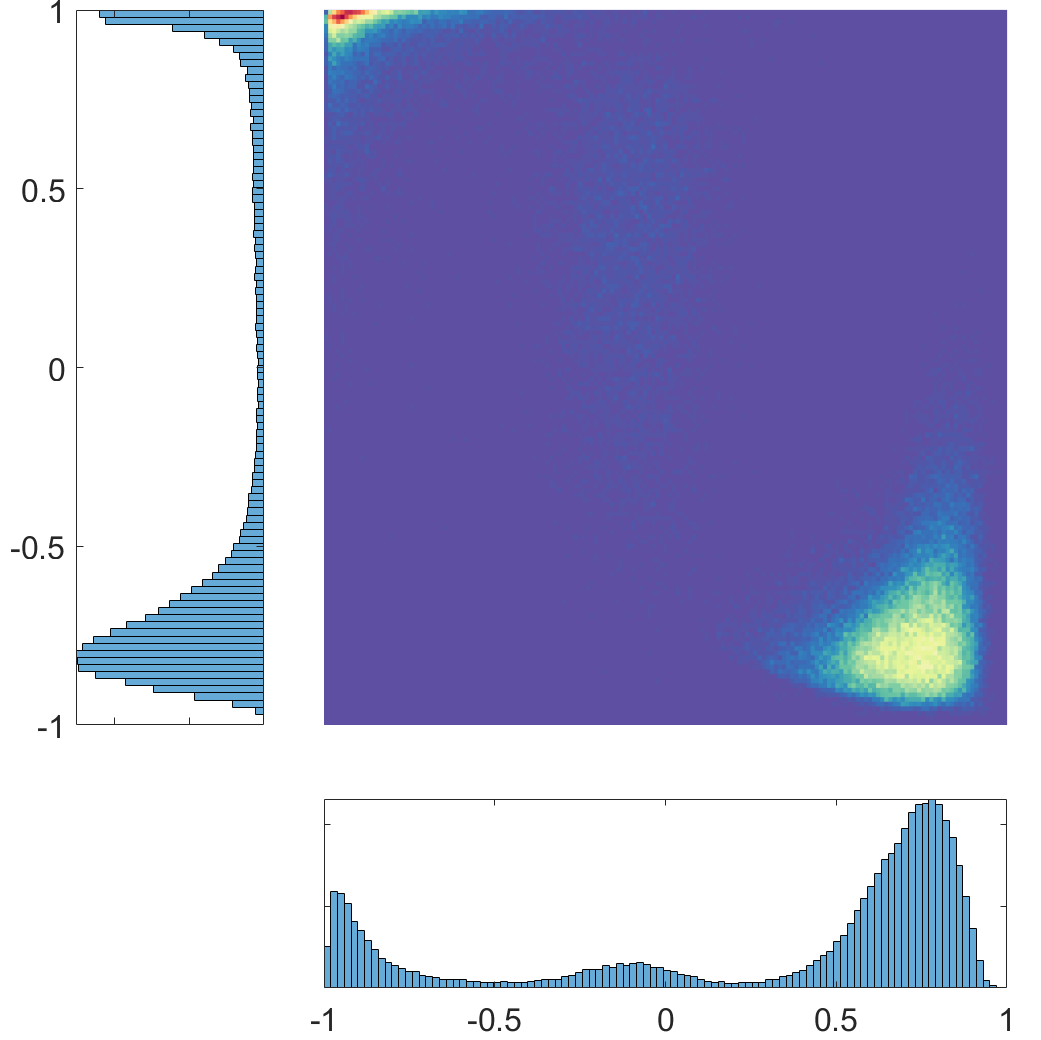} 
	}    
	\subfigure[Final phase]{
		\includegraphics[width=0.143\textwidth,height=0.11\textwidth]{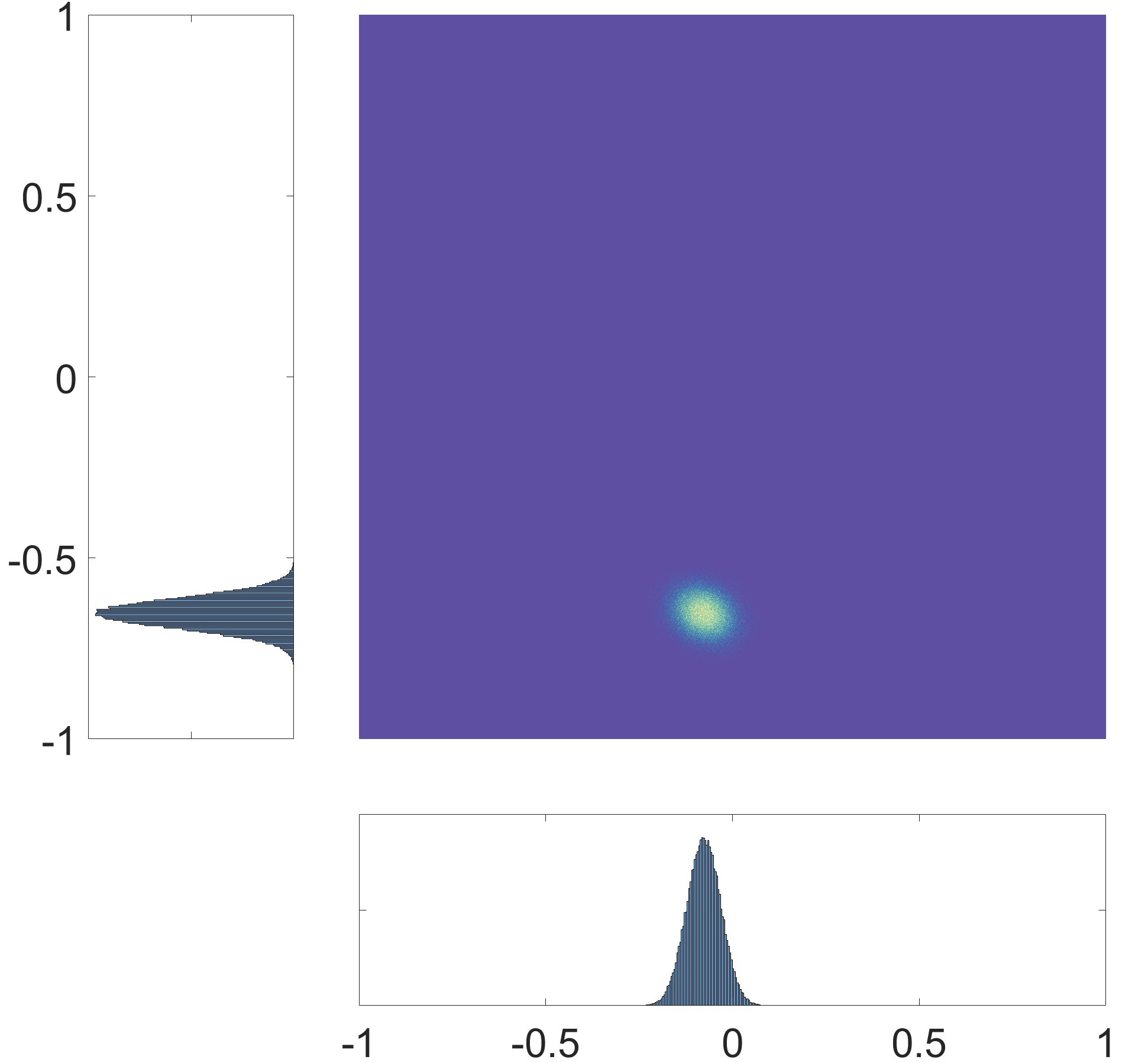} 
	}    
	\\
	\subfigure[Early phase]{
		\includegraphics[width=0.143\textwidth,height=0.11\textwidth]{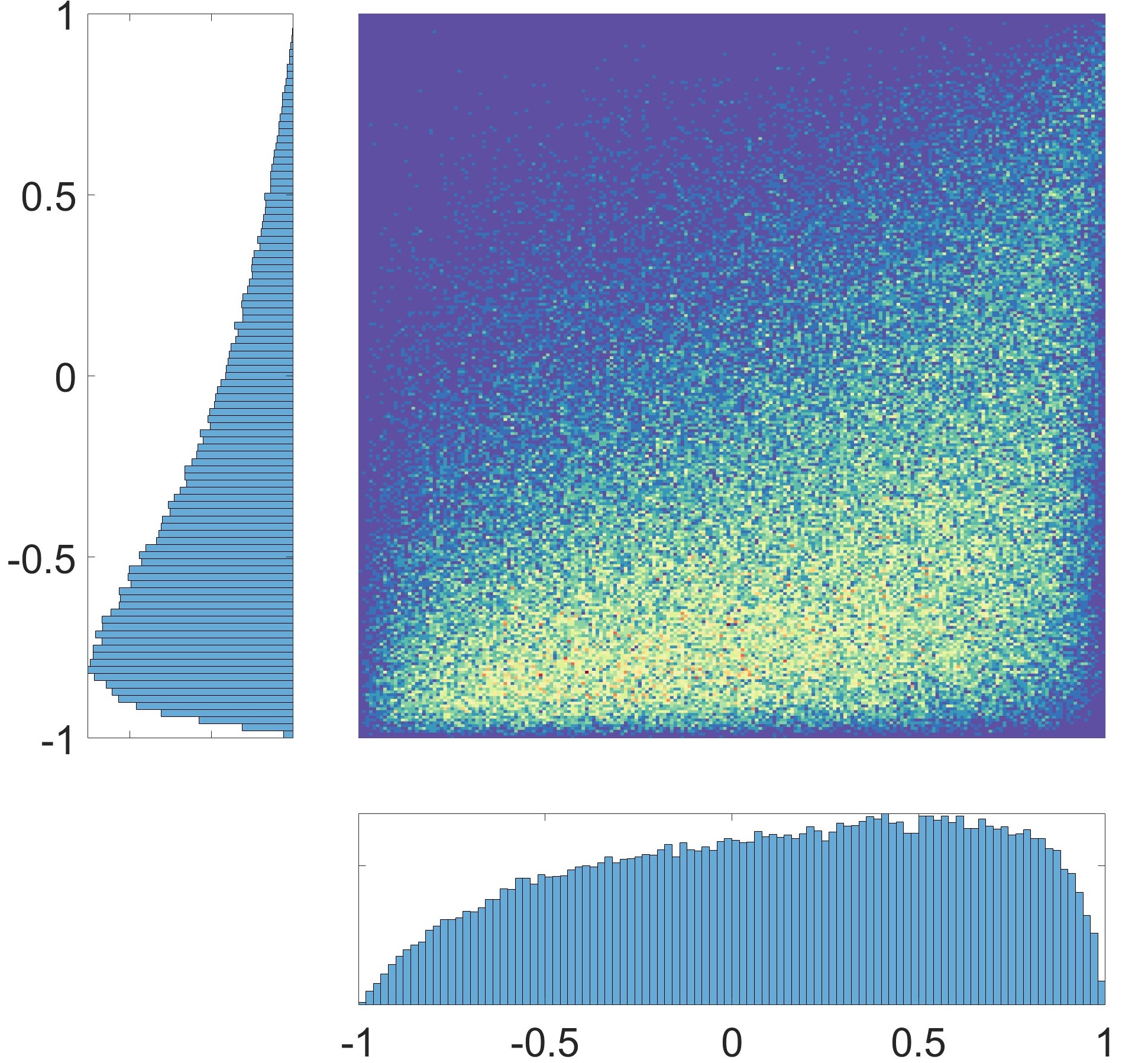} 
	}
	\subfigure[Middle phase]{
		\includegraphics[width=0.143\textwidth,height=0.11\textwidth]{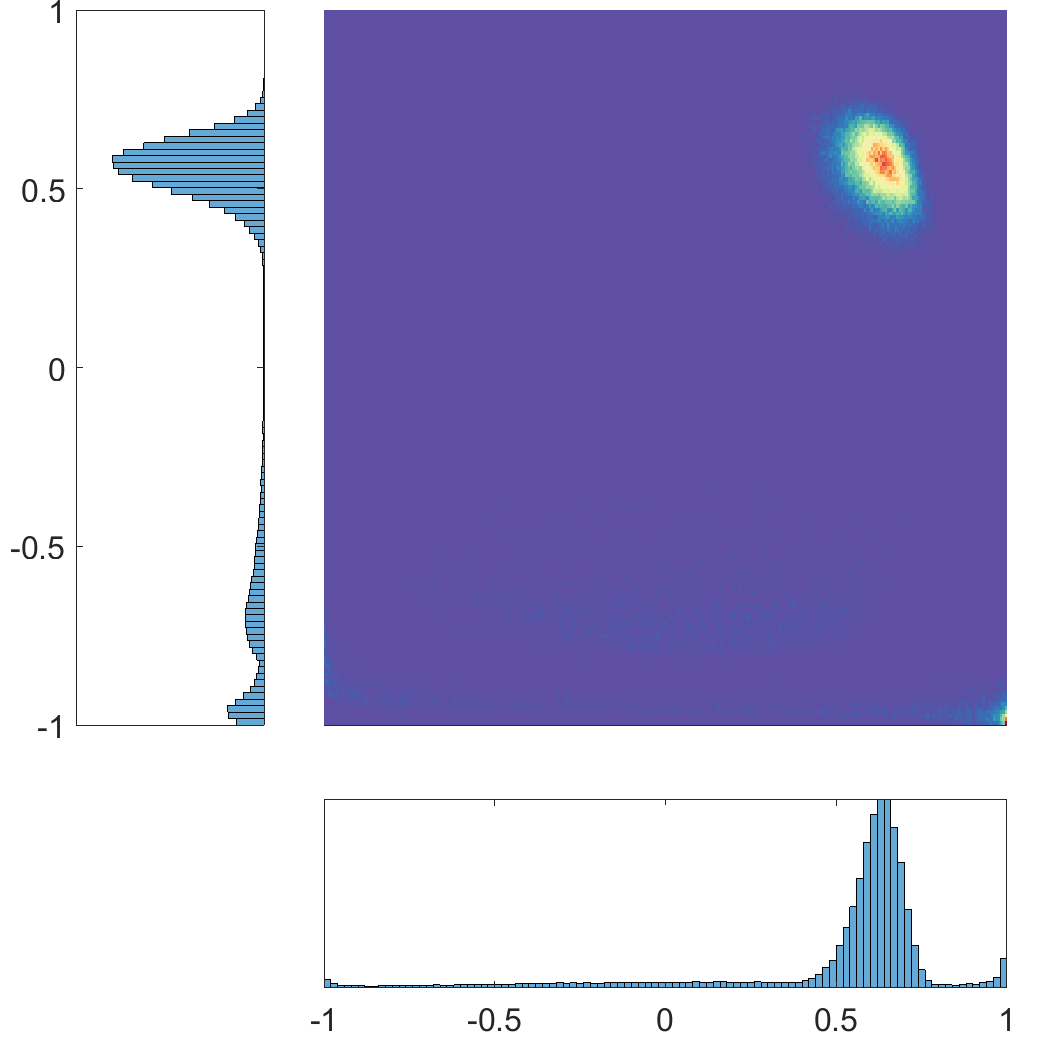} 
	}    
	\subfigure[Final phase]{
		\includegraphics[width=0.143\textwidth,height=0.11\textwidth]{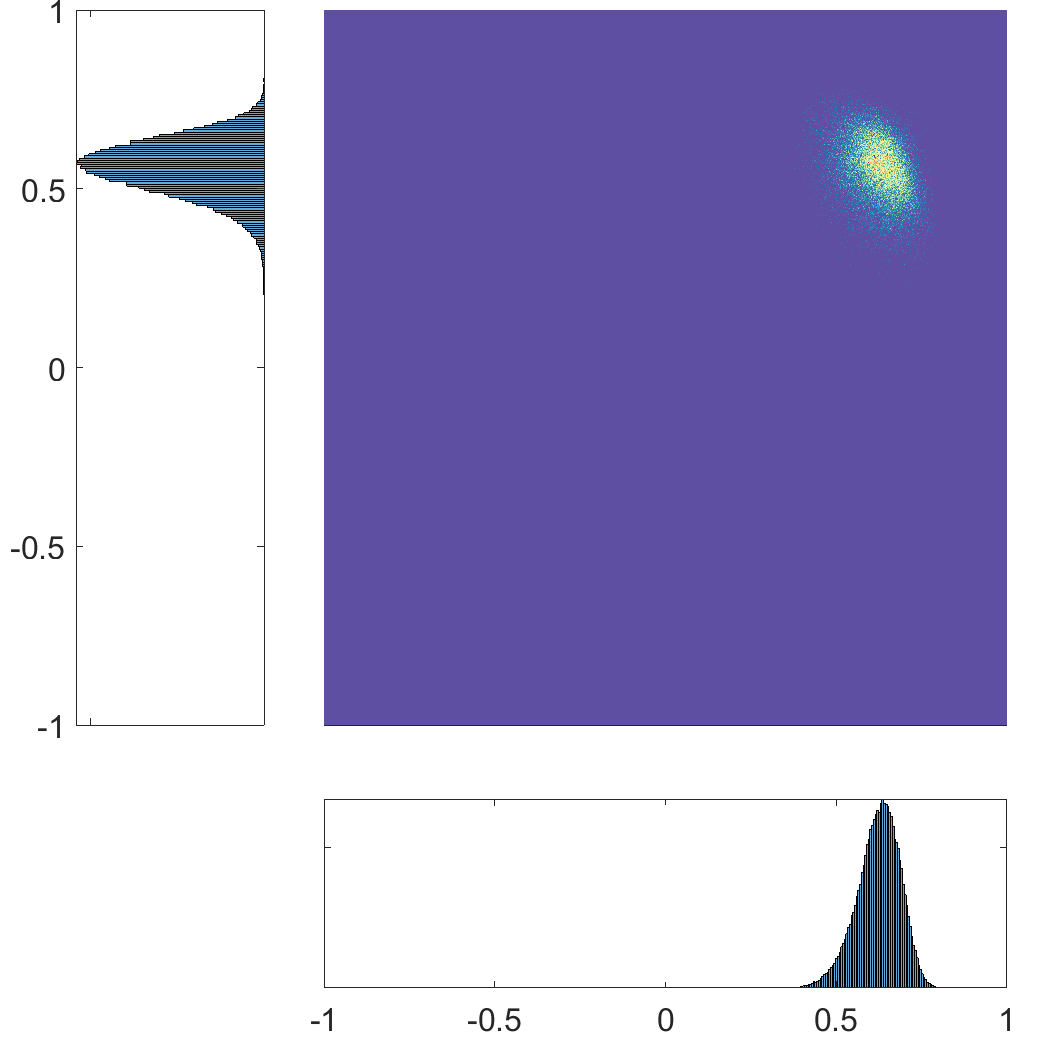} 
	}    
	\caption{Visualization of the action samples generated by push-forward policies in the early, middle and final phases.} \label{fig: heat map}
\end{figure}

\subsubsection*{(\romannumeral4) Exploration capability for push-forward policy}
We utilize customized navigation tasks to evaluate the exploration capability of the push-forward policy in PACER (see Fig. \ref{fig: custom env1}). In each test episode, the agent is required to navigate from a random initial position to a target zone while avoiding hazards and striving to collect treasures along the way. Whenever the agent reaches the target zone, it receives a reward, and the target zone is randomly refreshed until the maximum environmental interaction length is reached. The difficulty levels across tasks are controlled by the number of hazards and treasures set in the environments, and specific environment parameters are detailed in Appendix \ref{sec: appendix Experimental details}. As previously mentioned, the reward settings in the environment exhibit high randomness, necessitating that the policy thoroughly explores the environment before converging to the optimum, rather than becoming trapped in a local optimum during the early stages of training, such as pursuing only small rewards from treasures while neglecting the substantial rewards associated with reaching the target zone.

We compare the performance of PACER with other baselines on four such environments denoted by Env1 ... Env4.
We visualize PACER's push-forward policies at the early, middle, and final phase of the training process. Specifically, we sample 100000 actions for each state, and create a heat map on the two action dimensions in Fig. \ref{fig: heat map}. 
It is evident that the push-forward policies exhibit substantial exploration and multimodality modeling capabilities during the early phase and even the midst (500th over the 1000 total epochs) of PACER training, and would not degenerate into a deterministic policy fleetly. Simultaneously, it is capable of eventually converging to a relatively concentrating random policy near the optimal region, showcasing the exceptional performance of the PACER algorithm in striking a balance between exploration and exploitation.

Experiment results are shown in Table \ref{tab: table customized env} and Fig. \ref{fig: train curve full customized_envs}. From the results, it can be observed that both PACER$_M$ and PACER$_W$ achieved higher scores compared to other baselines, with PACER$_M$ outperforming all algorithms and achieving SOTA performance. Compared to the DAC algorithms, we find that expectation-based AC algorithms (SAC, TD3) are not well-suited for environments with high reward randomness, and even exhibit performance degradation during the training process (Fig. \ref{fig: train curve full customized_envs}). This could be attributed to the difficulty in convergence of the estimated value function $Q$ due to reward randomness, while the DAC algorithms, which model the return distribution $Z$, did not exhibit performance degradation.
This observation suggests that even with the use of entropy regularizer and constrained push-forward policy in SAC, the absence of a push-forward operator in the critic hinders its ability to perform well in environments with high randomness. 
The DSAC algorithm, which employs a push-forward critic and constrained push-forward policy, demonstrated improved performance compared to SAC, but still fell short of the PACER algorithm, which can fully leverage the modeling capabilities of the push-forward operator both in actor and critic. Based on the evidence presented, we can draw the clear conclusion that only by adopting highly expressive push-forward operators in both the actor and critic components can the procedure ignite an enhanced performance.


\begin{table}[htb]
	\centering
	\caption{Max average returns on the customized navigation tasks with $\pm$ 1 std over 7 different random seeds.} \label{tab: table customized env}
	\begin{tabular}{lcccc}
		\toprule
		& Env1  & Env2 & Env3 & Env4 \\
		\midrule
		\small PACER$_M$ & \textbf{599 $\pm$ 26}  &  \textbf{444 $\pm$ 10}  &  \textbf{531 $\pm$ 28} &  \textbf{477 $\pm$ 53}\\
		\small PACER$_W$ &  511$ \pm$ 38  &   400 $\pm$ 65  &  500 $\pm$ 33 &  460 $\pm$ 40\\
		\small IDAC & 549 $\pm$ 16   &  360 $\pm$ 117  &  439 $\pm$ 97 & 355 $\pm$ 65 \\
		\small DSAC & 494 $\pm$ 37   & 327 $\pm$ 116  & 423 $\pm$ 78 & 468 $\pm$ 36\\
		\small SAC & 323 $\pm$ 28  & 171 $\pm$ 113  & 164 $\pm$ 218 & 156 $\pm$ 118\\
		\small TD3 & 144 $\pm$ 49  & 191 $\pm$ 48  & 114 $\pm$ 68 & 107 $\pm$ 77\\
		\bottomrule
	\end{tabular}\small 
\end{table}

\section{Conclusions} \label{sec: conclusion}
We present PACER, the first fully push-forward-based DRL algorithm, in this paper. PACER leverages the push-forward operator to simultaneously model return distributions and stochastic policies, enabling the actor and critic with equal modeling capability, thus enhancing synergetic performance. To be compatible with the push-forward policy in PACER, we design a novel sample-based regularizer for efficient exploration and establish a stochastic utility value policy gradient theorem for policy update.
We validate the critical role of each component in our algorithm with extensive empirical studies. Experimental results demonstrate that PACER is capable of achieving SOTA performance on a number of challenging continuous control benchmarks.


\bibliographystyle{cas-model2-names}

\bibliography{cas-refs}





\clearpage
\onecolumn
\appendix
\section*{Appendix}
\section{Theoretical proof} \label{sec: appendix Theoretical study}

\subsection{Proof for Theorem \ref{th: utility bellamn}}
The proof of the following variant of the Bellman equation is analogous to the proof of the original Bellman equation.
\begin{equation}
    Z_{\psi}^{\pi} (s,a) \overset{\mathcal{D}}{=} \psi(R(s,a)) + \gamma Z_{\psi}^{\pi}(S',A')
\end{equation}
\begin{equation}
    Q_{\psi}^{\pi}(s,a) = \mathbb{E}_{R}[\psi(R(s,a))] + \gamma \mathbb{E}_{s' \sim P_s} [V_{\psi}^\pi(s')].
\end{equation}

\begin{proof}
    By definition, for the first equation:
    $$
    \begin{aligned}
    Z_{\psi}^{\pi}(s,a) &:= \sum_{t=0}^{\infty} \gamma^t \psi(R(s_t,a_t)) \\
    & \overset{\mathcal{D}}= \psi(R(s,a)) + \sum_{t=1}^{\infty} \gamma^t \psi(R(s_t,a_t)) \\
    & \overset{\mathcal{D}}= \psi(R(s,a)) + \gamma Z_{\psi}^{\pi}(S',A').
    \end{aligned}
    $$

    For the utility Bellman equation, let $S_{t+1:} = \{S_{t+1},S_{t+2},...\}$ and $A_{t+1:} = \{A_{t+1},A_{t+2},...\}$ denote the collection of all state and action random variables from time $t+1$ onwards. By definition,
    $$
    \begin{aligned}
    Q_{\psi}^{\pi}(s_t,a_t) &:=  \mathbb{E}_{S_{t+1:}, A_{t+1:} , R} Z_{\psi}^{\pi}(s_t,a_t)  \\
    & = \mathbb{E}_{S_{t+1:}, A_{t+1:} , R}\big[ \psi(R(s_t,a_t)) + \gamma Z_{\psi}^{\pi}(S_{t+1},A_{t+1}) \big] \\
    & = \mathbb{E}_{R}\psi(R(s_t,a_t)) + \gamma \mathbb{E}_{S_{t+1:}, A_{t+1:} , R} Z_{\psi}^{\pi}(S_{t+1},A_{t+1}) \\
    & = \mathbb{E}_{R}\psi(R(s_t,a_t)) + \gamma \mathbb{E}_{S_{t+1:}, A_{t+1:}}[ \mathbb{E}_{S_{t+2:}, A_{t+2:} , R} Z_{\psi}^{\pi}(S_{t+1},A_{t+1}) ] \\
    & = \mathbb{E}_{R}\psi(R(s_t,a_t)) + \gamma \mathbb{E}_{S_{t+1:}, A_{t+1:}} Q_{\psi}^{\pi}(S_{t+1},A_{t+1}) \\
    & = \mathbb{E}_{R}\psi(R(s_t,a_t)) + \gamma \mathbb{E}_{S_{t+1:}}V_{\psi}^{\pi}(S_{t+1})
    \end{aligned}
    $$
\end{proof}

\subsection{Proof for Theorem 2}
In this section, we present the proof of our main theorem. To ensure the validity of the proof in exchanging derivatives and integrals, as well as the order of integration, we consider the following common assumptions.

\begin{assumption}
    $P\left(s'| s, a\right), \nabla_a P\left(s' | s, a\right), \pi(\cdot|s,\xi,\theta_\pi), \nabla_{\theta_\pi} \pi(\cdot|s,\xi,\theta_\pi), R(s,a), \nabla_a R(s,a), \mu_0(s)$ are continuous in all parameters and variables $s, a, s'$ and $x$.
\end{assumption}

\begin{assumption}
    There exists bounds $B$ and $L$ such that $\sup_s \mu_0(s)<B, \sup _{a, s, s'} P\left(s'| s, a\right)<B, \sup _{a, s} R(s, a)<B$, $\sup _{a, s, s'}\left\|\nabla_a P\left(s' | s, a\right)\right\|<L$, and $\sup _{a, s}\left\|\nabla_a R(s, a)\right\|<L$.
\end{assumption}

The assumptions guarantee that $V^{\pi_\theta}(s)$ and $\nabla_{\theta_\pi} V^{\pi_\theta}(s)$ are continuous functions of $\theta_\pi$ and $s$. Furthermore, the compactness of $\mathcal{S}$ implies that for any $\theta_\pi,||\nabla_{\theta_\pi} V^{\pi_\theta}(s)||, || \nabla_a Q^{\pi_\theta}(s, a)|_{a=\pi(s,\xi,\theta_\pi)}||$ and $||\nabla_{\theta_\pi} \pi(s,\xi,\theta_\pi)||$ are bounded functions of $s$. 

\begin{theorem}[Stochastic Utility Value Policy Gradient (SUVPG)]\label{th: Stochastic utility Value Policy Gradient}
    For a push-forward policy $\pi(s,\xi;\theta_\pi)$ and a differentiable utility function $\psi(\cdot)$, the policy gradient of the state utility function $\mathbb{E}_{s \sim \mu_0}V_{\psi}^{\pi_{\theta}}(s)$ is given by
    \begin{equation}\label{eq: utility v target gradient}
        \nabla_{\theta_\pi} \mathbb{E}_{s \sim \mu_0}V_{\psi}^{\pi_{\theta}}(s) = 
        \mathbb{E}_{s \sim d_{\mu 0}^\pi, \xi \sim \mathscr{P}(\mathcal{X})}\left[\nabla_{\theta_\pi} \pi\left(s, \xi ; \theta_\pi\right) \cdot \nabla_a Q_{\psi}^{\pi_{\theta}}\left(s, a \right)\vert_{a = \pi(s, \xi;\theta_\pi)}\right].
    \end{equation}
\end{theorem}

\begin{proof}
    Firstly, let us remind that 
    \begin{equation}
        \begin{aligned}
        Q_{\psi}^{\pi}(s,a) := & \mathbb{E}_{s_{t+1} \sim P_s, a_{t+1} \sim\pi(\cdot|s_{t+1})}\left[\sum_{t=0}^{\infty}\gamma^t\psi(R(s_t,a_t))\vert_{s_0=s,a_0=a}\right], \\
        V_{\psi}^{\pi}(s) := & \mathbb{E}_{a \sim \pi(\cdot|s)}[Q_{\psi}^{\pi}(s,a)], \\
        Q_{\psi}^{\pi}(s,a)  = & \mathbb{E}_{R}[\psi(R(s,a))] + \gamma \mathbb{E}_{s' \sim P_s} [V_{\psi}^\pi(s')].
        \end{aligned}
    \end{equation}
    According to the definition of $J_{\psi}\left(\theta_\pi\right)$, its gradient can be written as  
    \begin{equation} \label{eq: target j psi}
        \nabla_{\theta_\pi}J_{\psi}(\theta_\pi) = \nabla_{\theta_\pi}\mathbb{E}_{s \sim \mu_0}V_{\psi}^{\pi_{\theta}}(s) 
        = \int_{\mathcal{S}} \mu_0(s) \nabla_{\theta_\pi} V_{\psi}^{\pi_{\theta}}(s) \mathrm{d}s.
    \end{equation}
    From here, we focus on the gradient of $V_{\psi}^{\pi_{\theta}}(s)$, and 
    \begin{equation} \label{eq: gradient v psi}
        \nabla_{\theta_\pi} V_{\psi}^{\pi_{\theta}}(s) = \nabla_{\theta_\pi} \int_{\mathcal{X}} P(\xi) Q_{\psi}^{\pi_{\theta}}\left(s, \pi\left(s, \xi ; \theta_\pi\right) \right) \mathrm{d}\xi 
        =  \int_{\mathcal{X}} P(\xi) \nabla_{\theta_\pi}Q_{\psi}^{\pi_{\theta}}\left(s, \pi\left(s, \xi ; \theta_\pi\right) \right) \mathrm{d}\xi.
    \end{equation}
    The gradient of $Q_{\psi}^{\pi_{\theta}}\left(s, \pi\left(s, \xi ; \theta_\pi\right) \right)$ in above equation can be calculated by
    \begin{equation*}
    \begin{aligned}
        \nabla_{\theta_\pi}Q_{\psi}^{\pi_{\theta}}\left(s, \pi\left(s, \xi ; \theta_\pi\right) \right) 
        &= \nabla_{\theta_\pi} [\psi\big(R(s,\pi\left(s, \xi ; \theta_\pi\right))\big) + \gamma \mathbb{E}_{s' \sim P_s(\cdot|s,\pi\left(s, \xi ; \theta_\pi\right))}V_{\psi}^{\pi_{\theta}}(s')] \\
        &= \nabla_{\theta_\pi} \psi\big(R(s,\pi\left(s, \xi ; \theta_\pi\right))\big) + \gamma \nabla_{\theta_\pi} \int_{\mathcal{S}} P_s(s'|s,\pi\left(s, \xi ; \theta_\pi\right))V_{\psi}^{\pi_{\theta}}(s')\mathrm{d}s' \\
        &= \nabla_{\theta_\pi} \pi\left(s, \xi ; \theta_\pi\right) \nabla_{a} \psi\big(R(s,a)\big)\vert_{a = \pi(s, \xi;\theta_\pi)} \\
        &\ \ + \int_{\mathcal{S}} \gamma \nabla_{\theta_\pi} \pi\left(s, \xi ; \theta_\pi\right) \nabla_{a} P_s(s'|s,\pi\left(s, \xi ; \theta_\pi\right))\vert_{a = \pi(s, \xi;\theta_\pi)} V_{\psi}^{\pi_{\theta}}(s')\mathrm{d}s' \\
        &\ \ + \int_{\mathcal{S}} \gamma P_s(s'|s,\pi\left(s, \xi ; \theta_\pi\right)) \nabla_{\theta_\pi} V_{\psi}^{\pi_{\theta}}(s')\mathrm{d}s' \\
        &= \nabla_{\theta_\pi} \pi\left(s, \xi ; \theta_\pi\right) \nabla_{a} [\psi\big(R(s,a)\big) + \int_{\mathcal{S}} \gamma P_s(s'|s,\pi\left(s, \xi ; \theta_\pi\right)) V_{\psi}^{\pi_{\theta}}(s')\mathrm{d}s' ]\vert_{a = \pi(s, \xi;\theta_\pi)} \\
        &\ \ + \int_{\mathcal{S}} \gamma P_s(s'|s,\pi\left(s, \xi ; \theta_\pi\right)) \nabla_{\theta_\pi} V_{\psi}^{\pi_{\theta}}(s')\mathrm{d}s' \\
        &= \nabla_{\theta_\pi} \pi\left(s, \xi ; \theta_\pi\right) \nabla_{a} Q_{\psi}^{\pi_{\theta}}(s,a)\vert_{a = \pi(s, \xi;\theta_\pi)} + \int_{\mathcal{S}} \gamma P_s(s'|s,\pi\left(s, \xi ; \theta_\pi\right)) \nabla_{\theta_\pi} V_{\psi}^{\pi_{\theta}}(s')\mathrm{d}s'. 
    \end{aligned}
    \end{equation*}
    By substituting $\nabla_{\theta_\pi}Q_{\psi}^{\pi_{\theta}}\left(s, \pi\left(s, \xi ; \theta_\pi\right) \right)$ back into \eqref{eq: gradient v psi}, we have 
    \begin{equation*}
    \begin{aligned}    
    \nabla_{\theta_\pi} V_{\psi}^{\pi_{\theta}}(s) & = \int_{\mathcal{X}} P(\xi)   \nabla_{\theta_\pi} \pi\left(s, \xi ; \theta_\pi\right) \nabla_{a} Q_{\psi}^{\pi_{\theta}}(s,a)\vert_{a = \pi(s, \xi;\theta_\pi)} \mathrm{d}\xi  \\
    &\ \ + \int_{\mathcal{X}} P(\xi)   \int_{\mathcal{S}} \gamma P(s \rightarrow s',1,\pi_\theta) \nabla_{\theta_\pi} V_{\psi}^{\pi_{\theta}}(s')\mathrm{d}s' \mathrm{d}\xi ,
    \end{aligned}
    \end{equation*}
    where $P(s \rightarrow s',1,\pi_\theta)$ indicates the probability that $s$ transforms to $s'$ in one step with policy $\pi_\theta$. We can see that  $\nabla_{\theta_\pi} V_{\psi}^{\pi_{\theta}}(s)$ have an iteration property, thus we can obtain that 
    \begin{equation*}
        \begin{aligned}    
        \nabla_{\theta_\pi} V_{\psi}^{\pi_{\theta}}(s) 
        & = \mathbb{E}_{\xi \sim \mathscr{P}(\mathcal{X})} \nabla_{\theta_\pi} \pi\left(s, \xi ; \theta_\pi\right) \nabla_{a} Q_{\psi}^{\pi_{\theta}}(s,a)\vert_{a = \pi(s, \xi;\theta_\pi)}  \\
        &\ \ + \mathbb{E}_{\xi, \xi' \sim \mathscr{P}(\mathcal{X})}  \int_{\mathcal{S}} \gamma P(s \rightarrow s',1,\pi_\theta) \nabla_{\theta_\pi} \pi(s', \xi';\theta_\pi)  \nabla_{a'} Q_{\psi}^{\pi_{\theta}}(s',a')\vert_{a' = \pi(s', \xi';\theta_\pi)} \mathrm{d}s' \\
        &\ \ + \mathbb{E}_{\xi, \xi' \sim \mathscr{P}(\mathcal{X})}  \int_{\mathcal{S}} \gamma^2 P(s \rightarrow s',2,\pi_\theta) \nabla_{\theta_\pi} V_{\psi}^{\pi_{\theta}}(s')  \mathrm{d}s' \\
        & = \dots \\
        & = \mathbb{E}_{\xi, \xi', \dots \sim \mathscr{P}(\mathcal{X})} \int_{\mathcal{S}} \sum_{t=0}^{\infty} \gamma^t P(s \rightarrow s',t,\pi_\theta) \nabla_{\theta_\pi} \pi(s', \xi';\theta_\pi) \nabla_{a'} Q_{\psi}^{\pi_{\theta}}(s',a')\vert_{a' = \pi(s', \xi';\theta_\pi)}  \mathrm{d}s' \\
        & = \mathbb{E}_{\xi \sim \mathscr{P}(\mathcal{X})} \int_{\mathcal{S}} \sum_{t=0}^{\infty} \gamma^t P(s \rightarrow s',t,\pi_\theta) \nabla_{\theta_\pi} \pi(s', \xi';\theta_\pi) \nabla_{a'} Q_{\psi}^{\pi_{\theta}}(s',a')\vert_{a' = \pi(s', \xi';\theta_\pi)}  \mathrm{d}s'
        \end{aligned}
    \end{equation*}
    As a result, we can conclude that 
    \begin{equation*}
    \begin{aligned}
        \nabla_{\theta_\pi}J_{\psi}(\theta_\pi) 
        & = \int_{\mathcal{S}} \mu_0(s) \mathbb{E}_{\xi \sim \mathscr{P}(\mathcal{X})} \int_{\mathcal{S}} \sum_{t=0}^{\infty} \gamma^t P(s \rightarrow s',t,\pi_\theta) \nabla_{\theta_\pi} \pi(s', \xi';\theta_\pi) \nabla_{a'} Q_{\psi}^{\pi_{\theta}}(s',a')\vert_{a' = \pi(s', \xi';\theta_\pi)}  \mathrm{d}s' \mathrm{d}s \\
        & = \mathbb{E}_{\xi \sim \mathscr{P}(\mathcal{X})} \int_{\mathcal{S}} d_{\mu_0}^{\pi_\theta} (s) \nabla_{\theta_\pi} \pi(s, \xi;\theta_\pi) \nabla_{a} Q_{\psi}^{\pi_{\theta}}(s,a)\vert_{a = \pi(s, \xi;\theta_\pi)}  \mathrm{d}s \\
        & = \mathbb{E}_{s \sim d_{\mu 0}^\pi, \xi \sim \mathscr{P}(\mathcal{X})}\left[\nabla_{\theta_\pi} \pi\left(s, \xi ; \theta_\pi\right) \cdot \nabla_a Q_{\psi}^{\pi_{\theta}}\left(s, a \right)\vert_{a = \pi(s, \xi;\theta_\pi)}\right].
    \end{aligned}
    \end{equation*}    
\end{proof}

\subsection{Stochastic value policy gradient}
As mentioned in the methodology section, the SUVPG theorem can be specialized to more specific cases. For instance, one such case arises when the utility function $\psi$ is an identity map, and the value function is formulated in the traditional expectation form. Under this case, the policy gradient for the push-forward policy can be calculated in a similar way.

\begin{corollary}[Stochastic Value Policy Gradient]
    For a push-forward policy $\pi(s,\xi;\theta_\pi)$, the gradient of $J\left(\theta_\pi\right) = \mathbb{E}_{s \sim \mu_0}V^{\pi_{\theta}}(s)$ is given by
    \begin{equation}\label{eq: Q target gradient}
    \nabla_{\theta_\pi} J\left(\theta_\pi\right)=\mathbb{E}_{s \sim d_{\mu 0}^\pi, \xi \sim \mathscr{P}(\mathcal{X})}\left[\nabla_{\theta_\pi} \pi\left(s, \xi ; \theta_\pi\right) \cdot \nabla_a Q^{\pi_{\theta}}\left(s, a \right)\vert_{a = \pi(s, \xi;\theta_\pi)}\right].
    \end{equation}
\end{corollary}

Notice that this corollary is nontrivial, as it provides the policy gradient for adopting push-forward policy under classic RL setting.

\begin{proof}
    The proof follows the same strategy as SUVPG theorem, and similar assumptions are required.
    \begin{equation*}
    \begin{aligned}
        \nabla_{\theta_\pi}J(\theta_\pi) &= \nabla_{\theta_\pi}\mathbb{E}_{s \sim \mu_0}V^{\pi_{\theta}}(s) \\
        &= \nabla_{\theta_\pi} \int_{\mathcal{S}} \mu_0(s) V^{\pi_{\theta}}(s) \mathrm{d}s \\
        &= \int_{\mathcal{S}} \mu_0(s) \nabla_{\theta_\pi} V^{\pi_{\theta}}(s) \mathrm{d}s.
    \end{aligned}
    \end{equation*}
    \begin{equation*}
    \begin{aligned}
    \nabla_{\theta_\pi} V^{\pi_{\theta}}(s) &= \nabla_{\theta_\pi} \int_{\mathcal{X}} P(\xi) Q^{\pi_{\theta}}\left(s, \pi\left(s, \xi ; \theta_\pi\right) \right) \mathrm{d}\xi \\
    &=  \int_{\mathcal{X}} P(\xi) \nabla_{\theta_\pi}Q^{\pi_{\theta}}\left(s, \pi\left(s, \xi ; \theta_\pi\right) \right) \mathrm{d}\xi.
    \end{aligned}
    \end{equation*}
    The gradient of $Q^{\pi_{\theta}}\left(s, \pi\left(s, \xi ; \theta_\pi\right) \right)$ in above equation can be calculated by
    \begin{equation*}
    \begin{aligned}
    \nabla_{\theta_\pi}Q^{\pi_{\theta}}\left(s, \pi\left(s, \xi ; \theta_\pi\right) \right) 
    &= \nabla_{\theta_\pi} [r(s,\pi\left(s, \xi ; \theta_\pi\right)) + \gamma \mathbb{E}_{s' \sim P(\cdot|s,\pi\left(s, \xi ; \theta_\pi\right))}V^{\pi_{\theta}}(s')] \\
    &= \nabla_{\theta_\pi} r(s,\pi\left(s, \xi ; \theta_\pi\right)) + \gamma \nabla_{\theta_\pi} \int_{\mathcal{S}} P(s'|s,\pi\left(s, \xi ; \theta_\pi\right))V^{\pi_{\theta}}(s')\mathrm{d}s' \\
    &= \nabla_{\theta_\pi} \pi\left(s, \xi ; \theta_\pi\right) \nabla_{a} r(s,a)\vert_{a = \pi(s, \xi;\theta_\pi)} \\
    &\ \ + \int_{\mathcal{S}} \gamma \nabla_{\theta_\pi} \pi\left(s, \xi ; \theta_\pi\right) \nabla_{a} P(s'|s,\pi\left(s, \xi ; \theta_\pi\right))\vert_{a = \pi(s, \xi;\theta_\pi)} V^{\pi_{\theta}}(s')\mathrm{d}s' \\
    &\ \ + \int_{\mathcal{S}} \gamma P(s'|s,\pi\left(s, \xi ; \theta_\pi\right)) \nabla_{\theta_\pi} V^{\pi_{\theta}}(s')\mathrm{d}s' \\
    &= \nabla_{\theta_\pi} \pi\left(s, \xi ; \theta_\pi\right) \nabla_{a} [ r(s,a) + \int_{\mathcal{S}} \gamma P(s'|s,\pi\left(s, \xi ; \theta_\pi\right)) V^{\pi_{\theta}}(s')\mathrm{d}s' ]\vert_{a = \pi(s, \xi;\theta_\pi)} \\
    &\ \ + \int_{\mathcal{S}} \gamma P(s'|s,\pi\left(s, \xi ; \theta_\pi\right)) \nabla_{\theta_\pi} V^{\pi_{\theta}}(s')\mathrm{d}s' \\
    &= \nabla_{\theta_\pi} \pi\left(s, \xi ; \theta_\pi\right) \nabla_{a} Q^{\pi_{\theta}}(s,a)\vert_{a = \pi(s, \xi;\theta_\pi)} + \int_{\mathcal{S}} \gamma P(s'|s,\pi\left(s, \xi ; \theta_\pi\right)) \nabla_{\theta_\pi} V^{\pi_{\theta}}(s')\mathrm{d}s'. \\
    \end{aligned}
    \end{equation*}
    So we have
    \begin{equation*}
    \begin{aligned}    
    \nabla_{\theta_\pi} V^{\pi_{\theta}}(s) & = \int_{\mathcal{X}} P(\xi)   \nabla_{\theta_\pi} \pi\left(s, \xi ; \theta_\pi\right) \nabla_{a} Q^{\pi_{\theta}}(s,a)\vert_{a = \pi(s, \xi;\theta_\pi)} \mathrm{d}\xi  \\
    &\ \ + \int_{\mathcal{X}} P(\xi)   \int_{\mathcal{S}} \gamma P(s \rightarrow s',1,\pi_\theta) \nabla_{\theta_\pi} V^{\pi_{\theta}}(s')\mathrm{d}s' \mathrm{d}\xi.
    \end{aligned}
    \end{equation*}
    Similarly, iterate this formula,
    \begin{equation*}  
    \nabla_{\theta_\pi} V^{\pi_{\theta}}(s) 
    = \mathbb{E}_{\xi \sim \mathscr{P}(\mathcal{X})} \int_{\mathcal{S}} \sum_{t=0}^{\infty} \gamma^t P(s \rightarrow s',t,\pi_\theta) \nabla_{\theta_\pi} \pi(s', \xi';\theta_\pi) \nabla_{a'} Q^{\pi_{\theta}}(s',a')\vert_{a' = \pi(s', \xi';\theta_\pi)}  \mathrm{d}s'.
    \end{equation*}
    Substitute into the original objective function, and we can get
    \begin{equation*}
    \begin{aligned}  
    \nabla_{\theta_\pi}J(\theta_\pi) 
    & = \int_{\mathcal{S}} \mu_0(s) \mathbb{E}_{\xi \sim \mathscr{P}(\mathcal{X})} \int_{\mathcal{S}} \sum_{t=0}^{\infty} \gamma^t \mathscr{P}(s \rightarrow s',t,\pi_\theta) \nabla_{\theta_\pi} \pi(s', \xi';\theta_\pi) \nabla_{a'} Q^{\pi_{\theta}}(s',a')\vert_{a' = \pi(s', \xi';\theta_\pi)}  \mathrm{d}s' \mathrm{d}s \\
    & = \mathbb{E}_{\xi \sim \mathscr{P}(\mathcal{X})} \int_{\mathcal{S}} d_{\mu_0}^{\pi_\theta} (s) \nabla_{\theta_\pi} \pi(s, \xi;\theta_\pi) \nabla_{a} Q^{\pi_{\theta}}(s,a)\vert_{a = \pi(s, \xi;\theta_\pi)}  \mathrm{d}s \\
    & = \mathbb{E}_{s \sim d_{\mu 0}^\pi, \xi \sim \mathscr{P}(\mathcal{X})}\left[\nabla_{\theta_\pi} \pi\left(s, \xi ; \theta_\pi\right) \cdot \nabla_a Q^{\pi_{\theta}}\left(s, a \right)\vert_{a = \pi(s, \xi;\theta_\pi)}\right].
    \end{aligned}
    \end{equation*}
\end{proof}

\clearpage
\subsection{Implementation of SUVPG in TD3} \label{sec: appendix Implementation of SUVPG in TD3}
We provide an implementation of SUVPG in Actor Critic based algorithm, e.g. Twin Delayed DDPG (TD3), in this section.
In comparison to DDPG, TD3 incorporates three techniques: (1) clipped double Q-learning, (2) the introduction of noise in the action function, and (3) a reduction in the update frequency of both the policy and target networks. 

We consider a revised version of TD3 that incorporates utility functions and the push-forward policy, abbreviated as PTD3.
Then, the object of PTD3 is maximizing $J_{\psi}(\theta_\pi) = \mathbb{E}_{s \sim \mu_0} V_{\psi}^{\pi_{\theta}}(s)$, and $\psi(\cdot)$ is a differentiable utility function as PACER. The push-forward policy $a = \pi(s,\xi;\theta_\pi)$ is used to replace the stochastic Gaussian policy in TD3.
According to the SUVPG theorem, $\nabla_{\theta_\pi}J_{\psi}(\theta_\pi) = \mathbb{E}_{s \sim d_{\mu_0}^{\pi}, \xi \sim \mathscr{P}(\mathcal{X})}[ \nabla_{\theta_\pi} \pi(s,\xi;\theta_\pi)\nabla_{a}Q_{\psi}^{\pi_\theta}(s,a)|_{a=\pi(s,\xi;\theta_\pi)} ]$

Thus, the pseudocode of PTD3 is provided below.

\begin{algorithm}
    \caption{PTD3}
    \begin{algorithmic}[1]
        \Require initial policy parameters $\theta_\pi$, Q-function parameters $\theta_{Q1}, \theta_{Q2}$, replay buffer $\mathcal{B}$, and other hyperparameters
        \State Set target parameters equal to main parameters $\hat{\theta}_{\pi} \leftarrow \theta_\pi, \hat{\theta}_{Q1} \leftarrow \theta_{Q1}, \hat{\theta}_{Q2} \leftarrow \theta_{Q2}$
        \Repeat
            \State Observe state $s$ and select action $a=\pi(s,\xi;\theta_\pi)$, where $\xi \sim \mathscr{P}(\mathcal{X})$
            \State Execute $a$, and store transition $\left(s, a, r, s', d\right)$ in replay buffer $\mathcal{B}$
            \If{it's time to update}
                \For{n = 1 to N}
                    \State Sample a batch $B_n=\{\left(s, a, r, s'\right)_{m=1}^M\}$ from $\mathcal{B}$
                    \For{each $(s, a, r, s')$ in $B_n$}
                        \State $\triangleright$ TD learning.
                        \State $a'=\pi(s,\xi;\hat{\theta}_\pi)$, where $\xi \sim \mathscr{P}(\mathcal{X})$
                        \State $\delta_1 = \left[Q_{\psi}(s,a;\theta_{Q1}) -  r - \gamma \min\{ Q_{\psi}(s',a';\hat{\theta}_{Q1}),Q_{\psi}(s',a';\hat{\theta}_{Q1}) \}\right]^2$
                        \State $\delta_2 = \left[Q_{\psi}(s,a;\theta_{Q2}) -  r - \gamma \min\{ Q_{\psi}(s',a';\hat{\theta}_{Q2}),Q_{\psi}(s',a';\hat{\theta}_{Q2}) \}\right]^2$
                    \EndFor
                    \State $\triangleright$ Update Value networks.
                    \State $\mathcal{L}(\theta_{Qi}) = \frac{1}{|B_n|} \sum_{(s, a, r, s') \in B_n} \delta_i$,for $i=1,2$
                    \State $\theta_{Qi} = \theta_{Qi} - \alpha \nabla_{\theta_{Qi}}\mathcal{L}(\theta_{Qi})$,for $i=1,2$
                    \If{$n$ mod $policy\_delay$ $=0$}
                    \State $\triangleright$ Update policy network.
                    \For{each $(s, a, r, s')$ in $B_n$}
                        \State $\overline{a} = \pi(s,\xi;\theta_\pi)$, $\xi \sim \mathscr{P}(\mathcal{X})$
                        \State $\theta_\pi = \theta_\pi + \beta \frac{1}{|B_n|} \sum_{s \in B_n}\nabla_{\theta_\pi} \pi(s,\xi;\theta_\pi)\nabla_{\overline{a}}Q_{\psi}^{\pi_\theta}(s,\overline{a})$ 
                    \EndFor
                    \State $\triangleright$ Update target networks.
                    \State $\hat{\theta}_{Qi} = \rho \hat{\theta}_{Qi}+(1-\rho) \theta_{Qi}$, for $i=1,2$
                    \State $\hat{\theta}_{\pi} = \rho \hat{\theta}_{\pi}+(1-\rho) \theta_\pi$
                    \EndIf
                \EndFor
            \EndIf
        \Until{convergence}
    \end{algorithmic}
\end{algorithm}

Furthermore, we can modify the objective function of PTD3 by incorporating a sample-based encourager, thereby promoting policy exploration in the environment. That is,
$$  
    J(\theta_\pi) = \mathbb{E}_{s \sim \mu_0} V_{\psi}^{\pi_{\theta}}(s) - \alpha \mathbb{E}_{s \sim \mathcal{B}}D_e\left( \pi(\cdot|s;\theta_\pi) || u(\cdot|s) \right).
$$

Thus, the resulting new algorithm can be implemented by just transforming the policy network update procedure in PTD3 to
$$
    \theta_\pi = \theta_\pi + \beta \frac{1}{|B_n|} \sum_{s \in B_n} \left[\nabla_{\theta_\pi} \pi(s,\xi;\theta_\pi)\nabla_{\overline{a}}Q_{\psi}^{\pi_\theta}(s,\overline{a}) - \nabla_{\theta_\pi}D_e\left( \pi(\cdot|s;\theta_\pi) || u(\cdot|s) \right) \right].
$$

\clearpage
\subsection{PACER with risk-measure utilities} \label{app: PACER risk measure}


The risk-measure type utility functions can be interpreted as expected value of some utility function $U$, i.e., $\mathbb{E}Z[U (Z(s, a))]$. If the utility function $U$ is linear, the policy obtained under such risk measure is called risk-neutral. A policy maximizing a linear utility function is called risk-neutral, whereas concave or convex utility functions give rise to risk-averse or risk-seeking policies, respectively \cite{dabney2018implicit}. In general, a distortion expectation is a generalized expression of risk measure which is generated from the distortion function.

\begin{definition}
    Let $h:[0,1] \rightarrow[0,1]$ be a distortion function such that $h(0)=0, h(1)=1$ and non-decreasing. Given a probability space $(\Omega, \mathcal{F}, \mathbb{P})$ and a random variable $Z: \Omega \rightarrow \mathbb{R}$, the distorted expectation corresponding to a distortion function $\phi$ is defined by:
    $$
    \mathbb{E}^{\phi(\mathbb{P})}[Z]=\int_{-\infty}^{\infty} z \frac{\partial}{\partial z}\left(\phi \circ F_Z\right)(z) d z,
    $$
    where $F_Z$ is the cumulative distribution function of $Z$.
\end{definition}

In fact, non-decreasing property of $h$ makes it possible to distort the distribution of $Z$ while satisfying the fundamental property of CDF. It can be showed that any distorted expectation can be expressed as weighted averages of quantiles \cite{dhaene2012remarks}. In other words, generating a distortion risk measure is equivalent to choosing a reweighting distribution:
\begin{equation}
    \label{eq: reweighting distribution}
    \mathbb{E}^{\phi(\mathbb{P})}[Z] = \int_0^1 F_Z^{-1}(\tau) d \phi  (\tau)
\end{equation}

Based on the foregoing, we can derive the objective function of the PACER algorithm under the Risk measure type, akin to that of the main paper.

The return distribution $Z^{\pi}(s,a)$ is given by
\begin{equation}
    Z^{\pi}(s,a) := \sum_{t=0}^{\infty} \gamma^t R(s_t,a_t),
\end{equation}
where $s_0=s,a_0=a,a_{t+1} \sim \pi(\cdot|s_{t+1})$. Remind the $Z^{\pi}(s,a)$ is modeled by IQN, i.e. $F_{Z(s,a)}^{-1}(\tau) = z(s,a,\tau;\theta_z)$.
$Q^{\phi,\pi}(s,a)$ can be defined as the distorted expectation of $Z^{\pi}(s,a)$:
\begin{equation} \label{eq: state-action risk-measure}
    Q^{\phi,\pi}(s,a) := \mathbb{E}_{s_{t+1} \sim P_s, a_{t+1} \sim \pi(\cdot|s_{t+1}), R}^{\phi(\mathbb{P})}Z^{\pi}(s,a),
\end{equation}
where $s_0=s,a_0=a$. 
Accordingly, $V^{\phi,\pi}(s)$ is defined by
\begin{equation}
    V^{\phi,\pi}(s) := \mathbb{E}_{a \sim \pi(\cdot|s)}[Q^{\phi,\pi}(s,a)].
\end{equation}

\begin{figure}[!t] 
	\centering
	\includegraphics[width=0.48\textwidth]{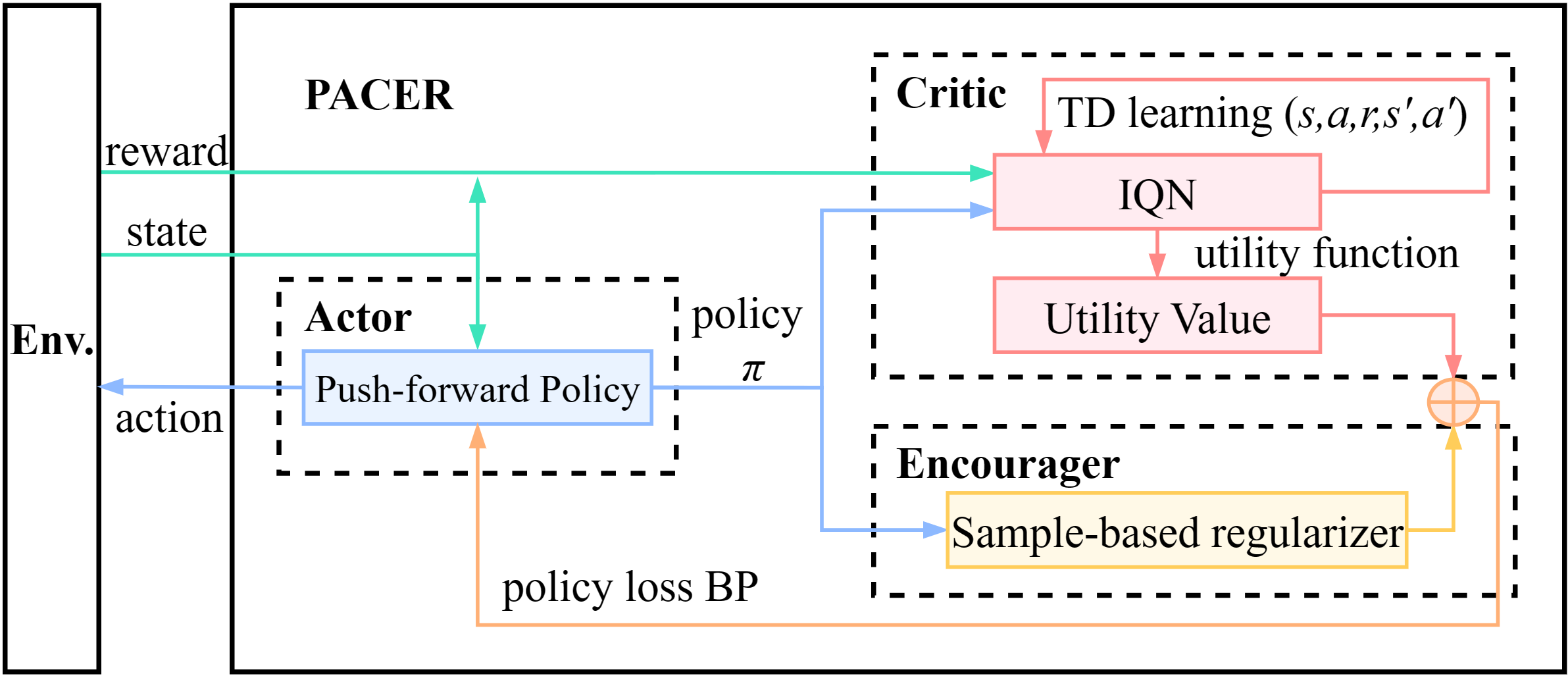} 
	\caption{The framework of PACER with risk-measure type utility.} \label{fig: diagram2}
\end{figure}

Similar to Eq. \ref{eq: ACE frame J}, the target function for PACER with risk-measure type utilities can be expressed as
\begin{equation}\label{eq: ACE frame J risk2}
	J^{\phi}\left(\theta_\pi\right) = \mathbb{E}_{s \sim \mu_0}V^{\phi,\pi_{\theta}}(s) - \alpha\mathbb{E}_{s \sim \mathcal{B}} D_e \big(\pi(\cdot|s;\theta_\pi)||u(\cdot|s) \big),
\end{equation}
and the network update procedure follows the same idea as before. The pseudocode is given in Algorithm \ref{alg: PACER risk measure}, and its framework is shown in Fig. \ref{fig: diagram2}

\begin{small}
	\begin{algorithm}[!tb] 
		\caption{PACER with risk-measure type utility}\label{alg: PACER risk measure}
		\begin{algorithmic}[1]
			\Require environment $env$, replay buffer $\mathcal{B}$, number of quantiles K, batch size M, discounted factor $\gamma$, base distribution $\mathscr{P}(\mathcal{X})$, regularizer sample number $N_r$, risk-measure type utility function $\phi$, weight $\alpha$, learning rate $\beta$, etc.
			\State initialize policy networks $\theta_{\pi}$,  value networks $\theta_z$ and target networks $\hat{\theta}_z$
			\For{$n=1$ to $N$}
			\State sample a transition $\{(s_n, a_n, r_n, s'_n)\}$ from $env$
			\State $\mathcal{B} = \mathcal{B} \cup \{(s_n, a_n, r_n, s'_n)\}$
			\If{$\frac{n}{N_{update}} == 0 \  \& \  n\geq N_{explore} $}
			\State sample $\mathcal{B}_n = \{(s, a, r, s')_{m=1}^{M}\}$ from $\mathcal{B}$
			\For{each $(s, a, r, s')$ in $\mathcal{B}_n$}
			\State $\triangleright$ Distributional TD learning.
			\State generate quantiles $\{\tau_i\}_{i=1}^{K}$, $\{\tau_j\}_{j=1}^{K}$
			\State $\hat{\tau}_i = \frac{\tau_{i+1}+\tau_i}{2}$, $ \hat{\tau}_j = \frac{\tau_{j+1}+\tau_j}{2}$
			\For{$i=1$ to $K$, $j=1$ to $K$}
			\State sample $\xi \sim \mathscr{P}(\mathcal{X})$
			\State $\hat{z} = r(s,a)+\gamma z (s', \pi(s',\xi;\theta_\pi),\hat{\tau}_i; \hat{\theta}_z)$
			\State $\delta_{i j}(s,a)=\hat{z} - z(s,a,\hat{\tau}_j;\theta_z)$ 
			\EndFor
			\State $\mathcal{L}^m(\theta_z) = \sum_{i=1}^{K} \sum_{j=1}^{K} \rho_{\hat{\tau}_i}^\kappa\big(\delta_{i j}(s,a)\big)$ 
			\State $\triangleright$ Calculate the sample-based regularizer.
			\For{$i=1$ to $N_{r}$}
			\State sample $\{\xi_{i},\xi_{j}\} \sim \mathscr{P}(\mathcal{X})$
			\State sample $\{a_i,a_j\}$ from random policy
			\State $x_i = \pi(s', \xi_{i};\theta_\pi)$,$x_j =\pi(s', \xi_{j};\theta_\pi)$
			\EndFor
			\State calculate $d_e^m(\theta_\pi)$ according to \eqref{equ:mmd-def} or \eqref{eq: P-wass samplebased}
			\State $\triangleright$ Calculate the utility value function.
			\State $\vartriangle_{\tau_i} = \tau_{i+1}-\tau_i$
			\begin{scriptsize}
				\State $V^m_{\psi}(s) = \sum_{i=1}^{K}\vartriangle_{\tau_i} \phi'(\hat{\tau}_i)z\big(s,\pi(s,\xi;\theta_{\pi}),\hat{\tau}_i;\theta_z\big)$
			\end{scriptsize}
			\EndFor
			\State $\triangleright$ Construct loss functions.
			\State $\mathcal{L}(\theta_z) = \frac{1}{M} \sum_{m=1}^{M}\mathcal{L}^m(\theta_z)$
			\State $\mathcal{L}(\theta_\pi) = \frac{1}{M} \sum_{m=1}^{M}(-V^m_{\psi}(s) + \alpha d_e^m(\theta_\pi))$ 
			\State update $\theta_z,\ \theta_\pi$ according their losses
            \State $\hat{\theta}_z = \beta\hat{\theta}_z + (1-\beta)\theta_z$
			\EndIf
			\EndFor
			\State \textbf{return} policy $\pi$ 
		\end{algorithmic}
	\end{algorithm}
\end{small}

\clearpage
\section{Experimental details} \label{sec: appendix Experimental details}
\subsection{Implementation details for PACER}\label{app: experiments details}
The Hyper-parameters settings and Network structure are given in Table \ref{tab: Hyper-parameters} and Table \ref{tab: Hyper-parameters}.

We use the following techniques in MuJoCo environments for training stability, all of them are also applied to baselines for fair comparisons.
\begin{itemize}
    \item \textbf{Observation Normalization}: in MuJoCo environments, the observation ranges from $-\infty$ to $\infty$. We normalize the observations by $clip\big((s - \hat\mu_{s})/(\max(\hat \sigma_{s})), -5, 5\big)$,
    where $\hat\mu_{s}$ is the mean of observations and $\hat\sigma_{s}$ is the standard deviation of observations.
    \item \textbf{Reward Scaling}: the reward signal for the environment HumanoidStandup is too large, so we shrink it  to 0.05 times its original reward for numerical stability. Notice that the change only reacts on training period, all testing results are carried out by the original reward signals.
\end{itemize}

\begin{table*}[!htbp]
    \centering
    \caption{Hyper-parameters settings}
    \label{tab: Hyper-parameters}
    \begin{tabular}{c|c} 
    \toprule
    \textbf{Hyper-parameters} & \textbf{Value}\\
    \hline
    Base distribution $\mathscr{P}(\mathcal{X})$ & $\mathcal{N}(0, 1)$ \\
    Number of quantiles K & $64$ \\
    Policy network learning rate & $3 \times 10^{-4}$ \\
    (Quantile) Value network learning rate & $3 \times 10^{-4}$ \\
    Encourager weight & $1 \times 10^{-2}$ or $1 \times 10^{-3}$  \\
    Optimizer & Adam \\
    Replay Buffer Size & $10^6$\\
    Total environment interactions & $1 \times 10^6$ \\
    Batch Size & $400$\\
    Number of training steps per update & $50$ \\
    Regularizer sample numbers $m$ & $100$ \\
    Discounted factor $\gamma$ & $0.99$ \\
    \bottomrule
    \end{tabular}
\end{table*}

\begin{table*}[!hb]
    \centering
    \caption{Network structure}
    \begin{tabular}{c|c} 
    \toprule
    \textbf{Actor} & \textbf{Critic}\\
    \hline
        (state dim + epsilon dim, 400) & (state dim + act dim, 400) \\
        Relu & Relu \\
        (400, 400) & (400, 400) \\
        Relu & Relu \\
        (400, action dim) & (400, 1) \\ 
        Tanh & \\
    \bottomrule
    \end{tabular}
\end{table*}

\subsection{Customized environments details} \label{app: Customized environments details}

Our customized environments are constructed using Gym \cite{brockman2016openai} and Safety-Gym \cite{ray2019benchmarking} packages. The customized environments are 2-D planes that including the following contents: (\romannumeral1) immovable treasures and moving treasures with positive rewards, (\romannumeral2) hazard regions with negative rewards, (\romannumeral3) a target zone with positive reward (higher than rewards for treasures), and (\romannumeral4) an agent with Lidars that can detect the orientation of the aforementioned objects. The agent is trained to navigate in the environment and gain as many rewards as possible. The rewards provided by the treasures and hazards can be gained repeatedly, i.e., the agent undergoes iterative rewards acquisition by interacting with treasures; correspondingly, any movement of the agent within hazards incurs penalties. Importantly, bonuses for obtaining treasures and penalties for entering hazards both occur in probabilistic forms. Specifically, their reward distributions follow the Bernoulli distribution $\mathcal{B}_{P}$, in other words, once the agent reach a treasure or moving in a hazard, it has probability $P$ to gain the corresponding reward and probability $1-P$ to get zero reward. 
In addition, whenever the agent moves towards the target area, it receives rewards. Upon entering the target zone, an immediate greater reward is bestowed to the agent, followed by the refreshment of the target area. In contrast to the sparse feedback obtained from treasures, the rewards associated with moving towards the target area are dense. Furthermore, due to the substantially higher reward for reaching the target area compared to other treasures, the agent achieves higher scores only by consistently progressing towards the target. 

In summary, the design philosophy of this environment aims to encourage the agent to continuously move towards the target area, collecting rewards along the way while avoiding hazard regions. 
A desired policy avoids trapping the agent into solely collecting treasures while overlooking the true objective. Following this approach, we designed four environments with varying levels of difficulty, and the specific configurations are presented in Table \ref{tab: Customized environment details}.

\begin{table*}[!t]
    \centering
    \caption{Customized environment details}\label{tab: Customized environment details}
    \begin{tabular}{l|c|c|c|c}
        \toprule
        \textbf{Settings} & Env1 & Env2 & Env3 & Env4 \\
        \hline
        Agent & car & car & car & car \\
        Task & reaching target & reaching target & reaching target & reaching target \\
        Number of environment steps & 1000 & 1000 & 1000 & 1000 \\
        Placement limits & [-2, -2, 2, 2] & [-5, -5, 5, 5] & [-3, -3, 3, 3] & [-4, -4, 4, 4] \\
        Observe goal Lidar & True & True & True & True \\
        Observe hazards Lidar & True & True & True & True \\
        Observe goal distance & True & True & True & True \\
        Observe treasures Lidar & True & True & True & True \\
        Reward for moving to target & 2.0 & 2.0 & 2.0 & 2.0 \\
        Reward for reaching goal & 5.0 & 8.0 & 8.0 & 8.0 \\
        Lidar maximum distance & 5 & 8 & 7 & 8 \\
        Lidar number of bins & 16 & 16 & 16 & 16 \\
        Number of hazards & 1 & 3 & 2 & 3 \\
        Hazards size & 0.5 & 0.8 & 0.5 & 0.6 \\
        Hazards cost & 1.0 & 2.0 & 2.0 & 2.0 \\
        Number of immovable treasures & 2 & 4 & 3 & 3 \\
        Immovable treasure size & 0.2 & 0.3 & 0.2 & 0.3 \\
        Immovable treasure reward & 0.2 & 0.2 & 0.2 & 0.2 \\
        Number of moving treasures & 2 & 3 & 3 & 3 \\
        Moving treasure keepout & 0.5 & 0.8 & 0.6 & 0.8 \\
        Moving treasure travel radius& 0.4 & 0.8 & 0.6 & 0.6 \\
        Moving treasure reward & 0.5 & 0.5 & 0.5 & 0.5 \\
        reward distribution & $\mathcal{B}_{0.65}$ & $\mathcal{B}_{0.65}$ & $\mathcal{B}_{0.65}$ & $\mathcal{B}_{0.65}$ \\
        cost distribution & $\mathcal{B}_{0.65}$ & $\mathcal{B}_{0.65}$ & $\mathcal{B}_{0.65}$ & $\mathcal{B}_{0.65}$ \\
        \bottomrule
    \end{tabular}
\end{table*}

\section{Sample-based adaptive weight-adjustment mechanism} 
Inspired by the automating temperature adjustment mechanism for maximum entropy RL \cite{haarnoja2018soft}, we propose a heuristic adaptive mechanism to automatically adjust the encourager's weight parameter $\alpha$ (see Eq. \ref{eq: ACE frame J}). The adaptive algorithm employed by SAC relies on the comparison between the entropy of the current policy and a predefined entropy threshold. This approach also has the limitation, repeatedly mentioned in this article, of "requiring explicit knowledge of the policy density." To be compatible with our sampled-based encourager, we have implemented a purely sample-based mechanism that can adaptively adjust the weight of the encourager, thereby more effectively trading off exploration and exploitation.

By considering the MMD regularizer as a constraint, we can reformulate $\max J_{\psi}\left(\theta_\pi\right)$ as the following constrained optimization problem:
	\begin{equation}
		\label{eq:constrained policy target}
		\begin{aligned}
			\max_{\theta_\pi}&\  \mathbb{E}_{s \sim \mu_0, \xi \sim \mathscr{P}(\mathcal{X})} \scalebox{0.8}{$\displaystyle\int_{0}^{1}$} F_{Z(s,\pi(s,\xi;\theta_\pi))}^{-1}(\tau) d\psi(\tau), \\
			s.t.\ & \mathbb{E}_{s \sim \mathcal{B}} D_e \big(\pi(\cdot|s;\theta_\pi)||u(\cdot|s) \big) \leq \beta. 
		\end{aligned}
	\end{equation}
\noindent Using Lagrange multipliers, the optimization problem can be converted into
\begin{equation}
\begin{aligned}
\max_{\theta_\pi} \min_{\alpha \ge 0} f(\theta_\pi,\alpha) =&  \mathbb{E}_{s \sim \mu_0, \xi \sim \mathscr{P}(\mathcal{X})} \scalebox{0.8}{$\displaystyle\int_{0}^{1}$} F_{Z(s,\pi(s,\xi;\theta_\pi))}^{-1}(\tau) d\psi(\tau) \\ 
& + \alpha \left[\beta - \mathbb{E}_{s \sim \mathcal{B}} D_e \big(\pi(\cdot|s;\theta_\pi)||u(\cdot|s) \big) \right]. 
\end{aligned}
\end{equation}
The above problem can be optimized by iteratively solving the following two sub-problems: $max_{\theta_\pi} J_{\psi}\left(\theta_\pi\right)$ and $min_{\alpha \ge 0} \mathcal{L}(\alpha)$, where
\begin{equation}
\begin{aligned}
    J_{\psi}\left(\theta_\pi\right) &= \mathbb{E}_{s, \xi} \scalebox{0.8}{$\displaystyle\int_{0}^{1}$} F_{Z(s,\pi(s,\xi;\theta_\pi))}^{-1}(\tau) d\psi(\tau) - \alpha \mathbb{E}_{s \sim \mathcal{B}} D_e \big(\pi(\cdot|s;\theta_\pi)||u(\cdot|s) \big),\\
    \mathcal{L}(\alpha) &= \alpha\left[\beta - \mathbb{E}_{s \sim \mathcal{B}} D_e \big(\pi(\cdot|s;\theta_\pi)||u(\cdot|s) \big)\right]. 
\end{aligned}
\end{equation}
The constraint $\mathbb{E}_{s \sim \mathcal{B}} D_e \big(\pi(\cdot|s;\theta_\pi)||u(\cdot|s) \big) \leq \beta$ restricts the feasible policy space within the realm of the reference policy.
Yet, the optimal $\beta$ could be varied from different training environments, which still needs manual tuning.
Actually, an unsuitable $\beta$ would greatly deteriorate the  performance. 

To avoid this issue, we implement a novel mechanism to adaptively adjust $\alpha$ and $\beta$, thus achieving a better balance between \emph{exploration and exploitation}. 
Intuitively, the policy should progressively acquire knowledge during training, leading to a gradual increase in the impact of exploration. 
Thus, we define the following objective for $\beta$
\begin{equation}
	\label{equ:beta-loss}
	\mathcal{L}(\beta) = \beta[\mathrm{sign}(\alpha_{\max} - \alpha) + \mathrm{sign}(\alpha_{\min} - \alpha)].
\end{equation}
When $\beta$ is fixed, $\alpha$ will increase to counter the rising trend of the encourager during training. 
A high $\alpha$ indicates that the current training period requires a larger $\beta$ value, prompting the policy to increase its exploitation rate. 
Conversely, a low $\alpha$ suggests that the current training period has an excessive $\beta$ value, prompting the policy to enhance exploration by decreasing $\beta$. The adoption of this mechanism introduces robustness and stability into PACER algorithm by realizing a proper trade-off between exploration and exploitation. 

We can set the parameters \(\alpha_{\min}\) and \(\alpha_{\max}\) within a broad range, such as (0, 2), to ensure that the PACER algorithm achieves a certain level of performance, see Table \ref{tab: weight-adjustment mechanism ablation} for results. Nevertheless, the optimal values for \(\alpha_{\min}\) and \(\alpha_{\max}\) still necessitate manual fine-tuning for each specific task, which is why we chose not to include this mechanism in the main algorithm presented in the text. The significance of proposing this adaptive mechanism lies in its nature as a sample-based mechanism, aligning with the sample-based framework of our paper. Enhanced methods for weight adaptation, building upon this approach, could be explored further.

\begin{table*}[!htb]
	\centering
	\caption{Comparison of max average returns between PACER$_M$ with adaptive weight and PACER$_M$ on MuJoCo environments } \label{tab: weight-adjustment mechanism ablation}
	\begin{tabular}{lcccccc}
		\toprule
		& Ant & Walker2d & Humanstandup & Humanoid & HalfCheetah & Hopper\\ 
		\midrule
		PACER$_M$-adaptive & \textbf{6831.07} & \textbf{5750.94} & \textbf{236213.0} & 5988.29 & 10918.1 & \textbf{3772.11} \\ 
		PACER$_M$ & 6386  & 5500  & 198856  & \textbf{6094}  & \textbf{12062}  & 3452 \\
        \bottomrule
	\end{tabular}
\end{table*}

\end{document}